\UseRawInputEncoding
\pdfoutput=1

\documentclass[10pt,journal,compsoc]{IEEEtran}
\ifCLASSOPTIONcompsoc
  \usepackage[nocompress]{cite}
\else
  \usepackage{cite}
\fi
\ifCLASSINFOpdf
\else
\fi
\long\def\comment#1{}

\newfont{\bbb}{msbm10 scaled 700}

\newfont{\bb}{msbm10 scaled 1100}

\newcommand{\PP}{\mbox{\bb P}}
\newcommand{\RR}{\mbox{\bb R}}

\newcommand{\ZZ}{\mbox{\bb Z}}

\newcommand{\EE}{\mbox{\bb E}}

\newcommand{\gv}{{\bf g}}

\newcommand{\mv}{{\bf m}}

\newcommand{\uv}{{\bf u}}
\newcommand{\wv}{{\bf w}}
\newcommand{\vv}{{\bf v}}
\newcommand{\xv}{{\bf x}}

\newcommand{\Dc}{{\cal D}}

\newcommand{\Ic}{{\cal I}}

\newcommand{\Lc}{{\cal L}}

\newcommand{\Oc}{{\cal O}}

\newcommand{\Sc}{{\cal S}}

\newcommand{\thetav}{\hbox{\boldmath$\theta$}}

\newcommand{\eqdef}{\stackrel{\Delta}{=}}

\newcommand{\trasp}{{\sf T}}

\usepackage{array}
\usepackage{multirow}
\usepackage{enumerate}
\usepackage{makecell}
\usepackage{tabularx} 
\usepackage{algorithm,algcompatible}
\usepackage{lipsum}
\usepackage{bbm}
\usepackage{times}
\usepackage[cmex10]{amsmath}
\usepackage{algpseudocode}
\usepackage{epsfig,verbatim}
\usepackage{amssymb} 
\usepackage{graphics,subfigure}
\usepackage{xcolor}
\usepackage{graphicx}
\usepackage{tabularx,booktabs,hhline}
\usepackage{color, colortbl}
\definecolor{LightCyan}{rgb}{0.88,1,1}
\definecolor{lightgray}{gray}{0.95}
\usepackage{multirow}
\usepackage{enumitem}

\newtheorem{theorem}{Theorem}

\newtheorem{definition}{Definition}

\newtheorem{lemma}{Lemma}

\newtheorem{corollary}{Corollary}

\newcommand{\argmin}{\operatornamewithlimits{argmin}}

\begin{document}
%
\title{Tighter Regret Analysis and Optimization of Online Federated Learning}

%
%
%
%

\author{Dohyeok~Kwon,~\IEEEmembership{Student Member,~IEEE}, Jonghwan~Park,~\IEEEmembership{Student Member,~IEEE} and Songnam~Hong,~\IEEEmembership{Member,~IEEE} 
\IEEEcompsocitemizethanks{\IEEEcompsocthanksitem D. Kwon and S. Hong are with the Department of Electronic Engineering, Hanyang University, Seoul, 04763, Korea.\protect\\
E-mail: \{kwon1585,snhong\}@hanyang.ac.kr
\IEEEcompsocthanksitem J. Park is with the Department of Electrical Engineering, University of Southern California, CA, 90089, USA.\protect\\
E-mail: jonghwan@usc.edu



}
}

\IEEEtitleabstractindextext{%
\begin{abstract}
In federated learning (FL), it is commonly assumed that all data are placed at clients in the beginning of machine learning (ML) optimization (i.e., offline learning). However, in many real-world applications, it is expected to proceed in an online fashion. To this end, online FL (OFL) has been introduced, which aims at learning a sequence of global models from decentralized streaming data such that the so-called cumulative regret is minimized. Combining online gradient descent and model averaging, in this framework, FedOGD is constructed as the counterpart of FedSGD in FL. While it can enjoy an optimal sublinear regret, FedOGD suffers from heavy communication costs. In this paper, we present a communication-efficient method (named OFedIQ) by means of intermittent transmission (enabled by client subsampling and periodic transmission) and quantization. For the first time, we derive the regret bound that captures the impact of data-heterogeneity and the communication-efficient techniques. Through this, we efficiently optimize the parameters of OFedIQ such as sampling rate, transmission period, and quantization levels. Also, it is proved that the optimized OFedIQ can asymptotically achieve the performance of FedOGD while reducing the communication costs by $99\%$. Via experiments with real datasets, we demonstrate the effectiveness of the optimized OFedIQ.
\end{abstract}

\begin{IEEEkeywords}
Online learning, federated learning, distributed optimization, streaming learning, regret analysis.
\end{IEEEkeywords}}

\maketitle
\IEEEdisplaynontitleabstractindextext
\IEEEpeerreviewmaketitle

\IEEEraisesectionheading{\section{Introduction}\label{sec:intro}}

Federated learning (FL) is an emerging framework for distributed and privacy-preserving machine learning (ML), in which a central server coordinates ML optimization with a massive number of clients without directly sharing training data stored in the decentralized clients (i.e., local data) \cite{hard2018federated, yang2019federated,kairouz2021advances, mammen2021federated, li2022fedipr}. In FL, the clients train local models separately from local data and the central server constructs a global model only using the aggregated local models \cite{mcmahan2017communication}. Specifically, a global model is learned by performing the two steps iteratively: i) local model optimizations (usually via stochastic gradient descent (SGD)) at each client; ii) global model optimization (e.g., model averaging) at the central server. There are myriads of applications such as ML from wearable devices \cite{elayan2021sustainability}, location-based services \cite{ciftler2020federated}, and human activity recognition \cite{ouyang2021clusterfl}.

Recently, numerous FL methods have been proposed.
FedSGD was introduced in \cite{mcmahan2017communication}, where a global model is optimized via local stochastic gradient descent (SGD) at clients and model averaging at a central server. Also, in \cite{khaled2020tighter}, the convergence analysis of FedSGD was provided for both cases of identical and heterogeneous data. Incorporating client subsampling (or partial participation) and periodic averaging into FedSGD, the de facto optimization method (named FedAvg) was presented in \cite{mcmahan2017communication}, which has lower communication costs than FedSGD. In \cite{reisizadeh2020fedpaq}, a communication-efficient FL method (named FedPAQ) was proposed by integrating a stochastic quantization into FedAvg. To sophisticate a quantization method, vector quantizations suitable for FL were proposed in 
\cite{shlezinger2020uveqfed, oh2022fedvqcs, brinkrolf2021federated}. Also, an enhanced gradient-quantization technique, called Lazily Aggregated Quantized (LAQ) gradient, was presented in \cite{sun2020lazily}.
Interestingly, it was shown in \cite{zhao2018federated,karimireddy2020scaffold} that FedAvg suffers from the 
so-called `client-drift' when the degree of data heterogeneity is high (i.e., non-IID data), resulting in unstable and slower convergence. This problem becomes more serious as the fraction of participating clients becomes smaller. Many works have put much effort into addressing the above problem. SCAFFOLD \cite{karimireddy2020scaffold} and Variance Reduced Local-SGD (VRL-SGD) \cite{liang2019variance} employed extra control variates to accelerate the convergence by reducing the variance of stochastic gradients. FedProx \cite{li2020federated} added a proximal term to each local optimizations in order to suppress the discrepancy among the local models. FedNova \cite{wang2020tackling} normalized the magnitude of local updates across the clients in the network so that the model averaging less distracts the global loss.

\subsection{Related Works}

In most of the existing FL works, it is commonly assumed that features (or data) do not possess an associated inherent ordering (i.e., non-streaming data) and accordingly, all data samples are placed at clients in the beginning of ML optimization (i.e., offline learning). Existing FL methods, thus, have paid little attention to ML optimization from continuous streaming (or time-series) data. In many real-world applications, however, ML tasks are expected to be performed in an {\em online} fashion, wherein data samples are generated as a function of time and clients have to predict a label (or make a decision) upon receiving an incoming data. To handle such tasks, {\em online} FL (OFL) has been introduced in \cite{hong2021communication} by combining online learning (OL) (or streaming learning) \cite{bubeck2011introduction} and FL. The goal of OFL is to seek a sequence of global models (or functions) in real time from continuous streaming data arriving at decentralized clients 
\cite{hong2021communication, gogineni2022communication}. At each time step, an up-to-date function is used to predict the label of newly incoming (previously unseen) data. The major difference between OFL methods and the aforementioned FL methods(e.g., FedAvg \cite{mcmahan2017communication}, FedPAQ\cite{reisizadeh2020fedpaq}, SCAFFOLD \cite{karimireddy2020scaffold}, FedProx \cite{li2020federated}, and FedNova \cite{wang2020tackling}) is in the objective of local updates. While FL methods aim at seeking a {\em sole} global model that minimizes a global loss function, OFL methods aim at seeking a sequence of global models that minimizes the so-called cumulative regret (see Section~\ref{sec:pre}) for details). Because of such difference, the existing FL methods may not ensure an optimal sublinear regret (i.e., an asymptotic optimality) for OFL. This opens up the necessity of new researches for OFL.

Very recently in \cite{hong2021communication}, an OFL method based on random feature-based multiple kernels (named eM-KOFL) was presented and its asymptotic optimality was proved via martingale argument. This work was mainly focused on dealing with multiple kernels in a communication-efficient way. Especially when a single-kernel is considered, vanilla method (named FedOGD) was constructed and analyzed, which can be immediately applied to deep neural network (DNN)-based methods. Note that FedOGD can be regarded as the counterpart of FedSGD \cite{mcmahan2017communication} in FL. While it can enjoy an optimal sublinear regret, FedOGD suffers from heavy communication costs. A communication-efficient OFL method (named PSO-Fed) was proposed in \cite{gogineni2022communication}, where clients and a central server exchange a fraction of their model parameters: however, no regret analysis was conducted. These two methods employed online gradient descent (OGD) (which is commonly used for OL) as the underlying online optimization method to the clients. Leveraging online mirror descent (OMD) and periodic averaging, FedOMD was presented in \cite{mitra2021online}. Note that when a widely-used Euclidean distance is adopted as Bregman divergence, OMD is reduced to OGD  \cite{hazan2016introduction}. It was proved that FedOMD can enjoy an optimal subliner regret. To the best of our knowledge, there is no additional OFL method with an analytical performance guarantee. 

We emphasize that compared with FL, OFL researches are at an infancy level. In particular, the existing regret analyses have not considered the impact of the fundamental communication-efficient techniques (e.g., client subsampling, periodic transmission, and quantization), and data-heterogeneity over time and across clients. This is the motivation of our work since such analysis would be the stepping stone for the research of OFL.

\subsection{Contributions}\label{subsec:cont}

In this paper, OFL is investigated in terms of constructing a communication-efficient method and tighter regret analysis. We improve the communication-efficiency of the baseline method (FedOGD) via two fundamental techniques such as intermittent transmission of model updates and quantization of updated gradients (see Fig. 1). The former directly reduces the number of transmissions (or active communication channels) and the latter diminishes the number of transmitted bits per each active channel. The proposed method is referred to as {\bf O}nline {\bf Fed}erated learning method with {\bf I}ntermittent transmission and {\bf Q}uantization (OFedIQ). Herein, intermittent transmission is enabled by client subsampling (a.k.a. partial client participation) and periodic transmission (a.k.a. periodic averaging), which stand for the infrequent transmissions across clients and over time steps, respectively. Our major contributions are summarized as follows.
\begin{itemize}
\item We derive tighter theory for OFL through a novel regret analysis. Our analysis is more rigorous than the existing ones such as the regret analysis of centralized OL \cite{bubeck2011introduction,shen2019random,hong2020active} and the regret analyses of distributed and federated OL \cite{chae2021distributed, hong2021communication, shen2021distributed, mitra2021online}. This is because our regret bound can reflect for the first time the effects of data-heterogeneity and the communication-efficient techniques (e.g., client subsampling, periodic transmission, quantization) on learning accuracy.

\item Through our analysis, we prove that OFedIQ over $T$ time steps can enjoy an optimal sublinear regret $\Oc(\sqrt{T})$ for any dataset (up to a constant), where the constant is bigger as data-heterogeneity grows. Namely, OFedIQ has the same asymptotic performance as FedOGD (i.e., the performance-limit) while significantly reducing the communication costs (e.g., about $99\%$ reduction).



\item For practical non-asymptotic cases, it is required to optimize the key parameters of OFedIQ such as sampling rate, transmission period, and quantization levels. Leveraging our regret bound, we efficiently tackle such optimization problem, yielding almost closed-form solutions (see Algorithm 3). Given target communication costs, the proposed Algorithm 3 provides the optimal parameters of OFedIQ with respect to maximizing the asymptotic learning accuracy.

\item Via experiments with real datasets, we demonstrate the superiority of the optimized OFedIQ on online classification and regression tasks. It can almost achieve the performance of FedOGD (i.e., the performance limit of communication-efficient OFL methods) while reducing the communication costs by $99\%$. Also, our method can outperform the benchmark methods such as OFedAvg and FedOMD for all datasets. These results suggest the practicality of the optimized OFedIQ.
\end{itemize}

\subsection{Outline and Notations}
The remaining part of this paper is organized as follows. In Section~\ref{sec:pre}, we formally define the objective and learning process of OFL. In Section~\ref{sec:methods}, we introduce the vanilla method (named FedOGD) and enhance its communication-efficiency by means of intermittent transmission and quantization. The resulting method is named OFedIQ. A tighter regret analysis is conducted in Section~\ref{sec:TA} and on the basis of this analysis, we efficiently optimize the key parameters of OFedIQ. In Section~\ref{sec:Exp}, we verify the effectiveness of our algorithm via experiments with real datasets. Some concluding remarks are provided in Section~\ref{sec:con}.

{\bf Notations:} Bold lowercase letters denote the column vectors. For any vector $\wv$,  $\wv^{\trasp}$ is the transpose of $\wv$ and $\|\wv\|$ is the $\ell_2$-norm of $\wv$. Also, $\langle\cdot,\cdot\rangle$ and $\EE[\cdot]$ represent the inner product in Euclidean space and the expectation over an associated probability distribution, respectively. To simplify the notations, for any positive integers $T, L$, we let $[T]\eqdef\{1,...,T\}$ and  $[T]_{L} \eqdef \{t \in [T]: t \mod L = 0\} \subseteq [T]$. We define a projection  $\psi_L:[T] \rightarrow [T]_{L} + 1$ such that
\begin{align}
    \psi_L(t) &\eqdef \lfloor (t-1)/L \rfloor \times L + 1 \mbox{ for some } t \in [T]. \label{eq:notation2}
\end{align} 
Also, $k$ and $t$ are used to indicate the indices of clients and time step, respectively.

%
%
\section{Online Federated Learning (OFL)}\label{sec:pre}
We formally define an {\em online} federated Learning (OFL) \cite{hong2021communication, mitra2021online, gogineni2022communication}. Recall that in (conventional) FL, ML optimization is tackled under the assumption that each client has access to local data samples offline and such samples are generated statistically. Whereas, OFL is basically formed by combining online learning (OL) and FL. Compared to FL, thus, OFL has the following two major differences:
\begin{itemize}[leftmargin=*]
    \item {\em Sequential data acquisition:} At every time step, each client has access to newly incoming (previously unseen) data. As in OL \cite{bubeck2011introduction}, there is no statistical assumption on data samples. 
    The detailed description of data acquisition in OFL  will be provided in Section~\ref{subsec:SDA}.
    \item {\em Online optimization:} Following the principle of OL, the objective of OFL is to learn a sequence of global models $\wv_1,\wv_2,...,\wv_{T}$  in an online fashion such that the so-called cumulative regret is minimized (see Section~\ref{subsec:OO}). 
\end{itemize} Remarkably, due to the aforementioned differences, it is required to conduct a new theoretical analysis for OFL, even if similar techniques in FL are applied.
\subsection{Sequential Data Acquisition}\label{subsec:SDA}

We consider a distributed learning framework consisting of $K$ clients indexed by $k \in [K]$ and a central server. At time step $t$, the central server broadcasts the up-to-date global model (denoted by $\wv_{t}$) to the $K$ clients. Accordingly, our learned function at the current time step is defined as $f(\xv;{\wv}_{t})$, from which every client $k \in [K]$ can estimate the label of newly incoming (previously unseen) feature data $\xv_{k,t} \in\mathcal{X}\subseteq\mathbb{R}^N$ in real time. We next focus on the local operations which are performed in a distributed way. At time step $t$, each client $k \in [k]$ starts to update its local model once it receives the true label of $\xv_{k,t}$, denoted by $y_{k,t}\in\mathcal{Y}\subseteq\mathbb{R}$. Given the labeled data $(\xv_{k,t}, y_{k,t})$ and the latest global model $\wv_{t}$, 
the client $k$ optimizes the local model (denoted as $\mv_{k,t} \in\mathbb{R}^D$) and sends it back to the central server. 
From the aggregated local models $\{\mv_{k,t}: k \in [K]\}$, the central server constructs an updated global model by simply averaging them:
\begin{align}
        {\wv}_{t+1}=\frac{1}{K}\sum_{k=1}^{K}\mv_{k,t}, \label{eq:update_global}
\end{align}and then broadcasts it back to the $K$ clients.

\vspace{0.1cm}
\noindent {\bf Communication costs (per time step):} The communication costs of an OFL method are measured by the number of total transmitted bits of local information (e.g., $\mv_{k,t}$'s) conveyed from the $K$ clients to the central server (in short, uplink communication costs). For instance, if every client sends a $D$-dimensional real-valued vector (i.e., $\mv_{k,t}\in\RR^D$) to the central server, the corresponding communication costs are equal to $32KD$ bits. Herein, 32-bit quantization is typically assumed to represent a real-value precisely. The communication costs from the central server to the $K$ clients (i.e., downlink communication costs) are not of concern as they are relatively very small due to the broadcasting nature.

\subsection{Online Optimization}\label{subsec:OO}

We describe an OFL protocol focusing on the ML optimization of a sequence of global models.  At every time step $t=1,2,...,T$:
\begin{itemize}[leftmargin=*]
    \item A global model $\wv_{t}$ is updated by averaging the aggregated local models $\{\mv_{k,t-1}: k\in [K]\}$.
    \item Simultaneously a new local data $(\xv_{k,t},y_{k,t})$ is generated at each client $k$ for $\forall k \in [K]$.
    \item Each client $k\in[K]$ suffers from the local loss $\Lc(f(\xv_{k,t};\wv_{t}),y_{k,t})$, where $\Lc(\cdot,\cdot)$ is a loss (or cost) function that measures the error between true and predicted labels.
    \item Leveraging the local-loss, each client $k$ optimizes its local model $\mv_{k,t}$.
\end{itemize} As in (centralized) OL \cite{hazan2016introduction, bubeck2011introduction, hong2020active, hong2021communication}, the objective of OFL is to seek a sequence of global models $\wv_{1},\wv_{2},...,\wv_{T}$ that minimizes the {\em cumulative regret} over $T$ time steps:
\begin{align}
        {\rm regret}_{T} &= \sum_{t=1}^{T}\sum_{k=1}^{K}\Lc({f}(\xv_{k,t};{\wv}_{t}),y_{k,t}) \nonumber\\
        &\quad\quad\quad - \min_{\wv \in \RR^D} \sum_{t=1}^{T}\sum_{k=1}^{K} \Lc(f(\xv_{k,t};\wv), y_{k,t}).\label{eq:metric}
\end{align} The  upper  bound  on $\mbox{regret}_T$ is  referred  to as {\em regret bound}. This metric compares the cumulative loss of an OFL method with that of the {\em static} optimal function in hindsight. We emphasize that due to the sequential optimization nature of OFL protocol, its performance metric in \eqref{eq:metric} becomes totally different from that of (conventional) FL methods. To be specific, the local steps of FL methods aim at optimizing an objective function such as the expected global loss over local data distributions, whereas those of OFL methods aim at minimizing the cumulative regret in \eqref{eq:metric}. Therefore, it is required to conduct a new analysis for OFL methods even if they are constructed using the well-proven techniques in the existing FL methods. We contribute to this subject.


\section{Methods}\label{sec:methods}

In this section, we introduce a vanilla method for OFL as the counterpart of FedSGD for FL, and then improve its communication-efficiency by means of intermittent transmission and quantization.

\subsection{Vanilla Method: FedOGD}\label{subsec:vanilla}

We construct a vanilla method by integrating online gradient descent (OGD) \cite{hazan2016introduction} into OFL. This method is named FedOGD (i.e., the de facto algorithm in OFL) as the counterpart of FedSGD in FL  \cite{mcmahan2017communication}. To learn a sequence of global models in an online fashion, at every time step $t\in[T]$, FedOGD consists of the following two steps (see {\bf Algorithm 1}):

\vspace{0.1cm}
\noindent{\bf i) Local update:} At time step $t$, each client $k\in[K]$ is aware of a current global model ${\wv}_{t}$ (which is conveyed from the central server at time step $t-1$) and receives an incoming data $(\xv_{k,t},y_{k,t})$. Leveraging them, it optimizes the local model via OGD \cite{hazan2016introduction}:
\begin{equation}
    {\wv}_{k,t+1} = {\wv}_{t} - \eta \nabla\Lc\left(f(\xv_{k,t};{\wv}_{t}),y_{k,t} \right),\label{eq:local1}
\end{equation} with a step size (or learning rate) $\eta>0$ and the initial value ${\wv}_{1}$, where $\nabla\Lc\left(f(\xv_{k,t};{\wv}_{t}),y_{k,t} \right)$ is the gradient at the point ${\wv}_{t}$. Then, every client $k$ sends the updated local model $\mv_{k,t}={\wv}_{k,t+1}\in \RR^{D}$ back to the central server. The communication costs of FedOGD are equal to $32DK$ bits.

\vspace{0.1cm}
\noindent{\bf ii) Global update:} From the aggregated local models $\{\mv_{k,t}={\wv}_{k,t+1}: k\in[K]\}$, the central server constructs a global model via averaging:
\begin{align}
    {\wv}_{t+1} &= \frac{1}{K} \sum_{k=1}^{K} {\mv}_{[k,t]}\nonumber\\
    &=\wv_{t} - \frac{\eta}{K}\sum_{k=1}^{K}\nabla\Lc\left(f(\xv_{k,t};{\wv}_{t}),y_{k,t} \right). \label{eq:global1}
\end{align} Then, it broadcasts the updated global model ${\wv}_{t+1}\in\RR^{D}$ to all $K$ clients.


\begin{algorithm}
\caption{Vanilla Method: FedOGD}\label{alg:vanilla}
\begin{algorithmic}[1]
\State {\bf Input:} $K$ clients and learning rate $\eta$.
\State {\bf Output:} A sequence of function (or global) parameters $\{\wv_{t+1}:t\in [T]\}$.
\State {\bf Initialization:} $\wv_{1}$ and $\wv_{k,1}$ for $\forall k \in [K]$.
\State {\bf Iteration:} $t=1,2,\ldots,T$
    \Statex$\diamond$ At the client $k\in[K]$
        \Statex\quad$\bullet$ Receive streaming data $(\xv_{k,t},y_{k,t}).$
        \Statex\quad$\bullet$ Update local parameter $\wv_{k,t+1}$ via \eqref{eq:local1}.
        \Statex\quad$\bullet$ Transmit $\mv_{k,t}=\wv_{k,t+1}$ to the central server. 
    \Statex$\diamond$ At the central server
        \Statex\quad$\bullet$ Receive $\mv_{k,t}$ from the $K$ clients $k\in[K]$. 
        \Statex\quad$\bullet$ Update the global parameter $\wv_{t+1}$ via \eqref{eq:global1}.
        \Statex\quad$\bullet$ Broadcast $\wv_{t+1}$ to all $K$ clients.
\end{algorithmic}
\end{algorithm}

\begin{figure*}
\centerline{\includegraphics[width=15cm]{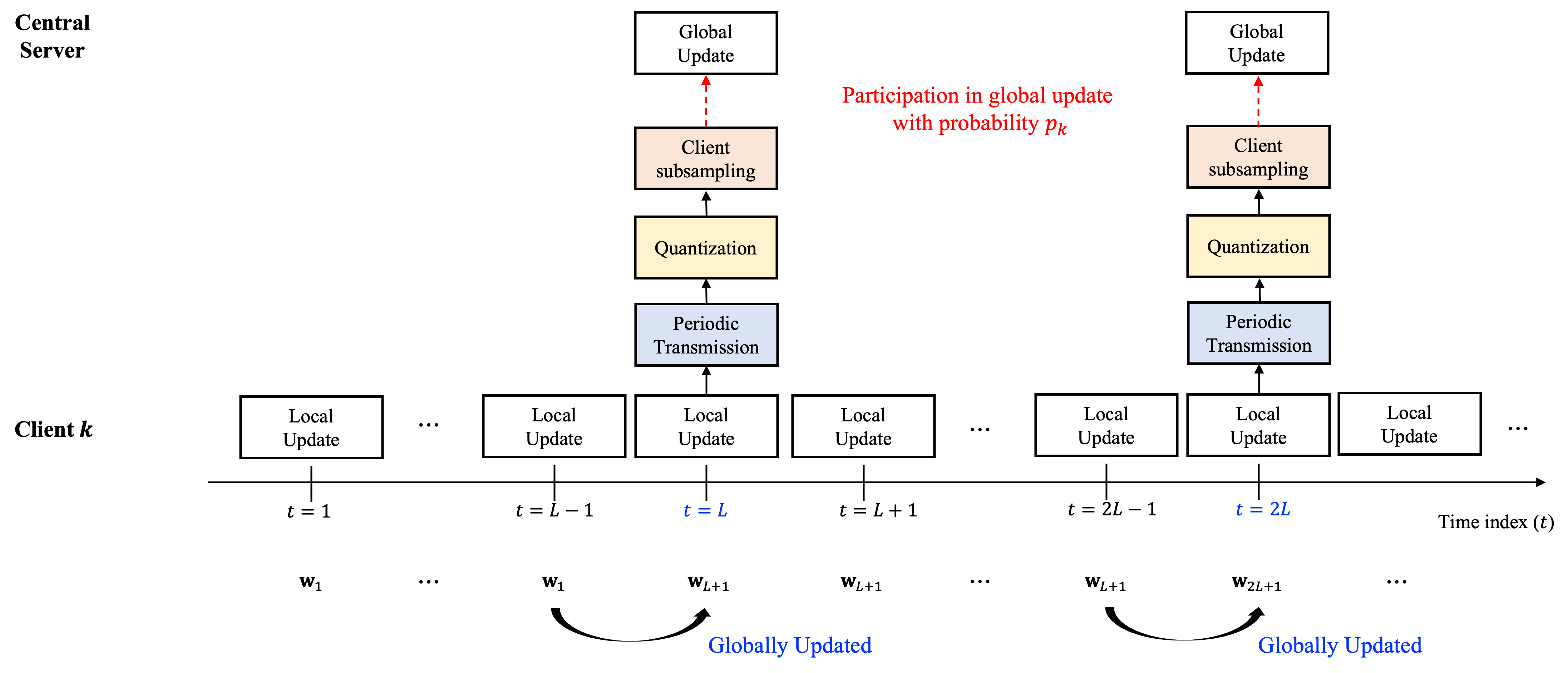}}
\caption{Description of the proposed OFedIQ for an uplink transmission between the client $k$ and the central server.}
\label{fig:description}
\end{figure*}

\subsection{Communication-Efficient Method: OFedIQ}\label{sec:methods}

To alleviate the heavy communication costs, numerous communication-reduction techniques have been proposed for conventional FL \cite{reisizadeh2020fedpaq, shlezinger2020uveqfed, oh2022fedvqcs, brinkrolf2021federated}. Among them, the key features can be summarized as follows: (i) client subsampling (a.k.a. partial client participation); ii) periodic transmission (a.k.a. periodic averaging); (iii) quantization. Note that (i) and (ii) enable intermittent transmission, thereby directly reducing the number of transmissions, and (iii) decreases the number of transmitted bits per communication-channel. We first describe the three key features in the context of OFL (see Fig.~\ref{fig:description}).

\vspace{0.1cm} 
\noindent{\bf Client subsampling:} In OFL, there are lots of clients (e.g., smart phone users) communicating with a central server via an uplink communication channel. At every time step, thus, it is impractical to have all clients participating in OFL protocol due to the extremely heavy communication costs. As in (conventional) FL methods \cite{mcmahan2017communication, reisizadeh2020fedpaq}, client subsampling, where a small fraction of clients are participated in OFL protocol at every time step, is necessary for OFL methods. To that end, we in this paper adopt the probabilistic client subsampling: each client $k \in [K]$ can be sampled (or participated) with probability $0 < p_k < 1$ independently at every time step and from other clients, where $p_k$ indicates the sampling rate (or participation probability). Note that various $p_k$'s can take into account system-heterogeneity since higher probabilities can be assigned to the clients with higher computation capabilities or communication capacities. We remark that this probabilistic approach is required for the compactness of tighter regret analysis while any approach can be immediately applied to the proposed OFL method. Throughout the paper, $\Sc_t \subseteq [K]$ represents the index subset of participating (or sampled) clients at time step $t$. In convention, full client participation implies that $p_k=1$ for $\forall k \in [K]$ (i.e., $\Sc_t=[K]$ for $\forall t$). Obviously, when $p_k < 1$ for some $k$, $\Sc_t \subseteq [K]$ is a random variable due to our probabilistic modeling, namely, a time-varying client subsampling is considered. When our client subsampling is applied, the communication cost can be reduced to $32D\sum_{k=1}^{K}p_k$ bits in average. In the homogeneous setting with $p_k=p$ for $\forall k$, they can be simplified as $32pKD$ bits in average


\vspace{0.1cm}
\noindent{\bf Periodic transmission:} Periodic transmission is one of the key techniques to reduce the communication costs in FL \cite{reisizadeh2020fedpaq}. However, in OFL, this method might degrade the performance (i.e., the cumulative regret in \eqref{eq:metric}) as a new global model $\wv_{t+1}$ should be constructed at every time step. This question will be addressed in Section~\ref{sec:TA} based on our tighter regret analysis.  
Assuming the transmission period $L\geq 1$, each client $k \in [K]$ sends the local information $\mv_{k,t}$ to the central server at every $L$ time steps. Due to the infrequent transmission, a sequence of learned global models is obtained as
\begin{equation*}
    \underbrace{\wv_{1}\rightarrow \cdots \rightarrow \wv_{1}}_{L} \rightarrow \underbrace{\wv_{L+1}\rightarrow \cdots \rightarrow \wv_{L+1}}_{L}\rightarrow \wv_{2L+1}\cdots.
\end{equation*} 
Thus, for some positive integer $a$, each client $k$ estimates the label of newly incoming data at time step $t \in [aL+1, (a+1)L+1)$ such as
\begin{equation*}
    \hat{y}_{k,t}=f(\xv_{k,t};\wv_{aL+1}).
\end{equation*} For example, during the $L$ time steps  $ 2L+1 \leq t \leq 3L$, the identical global model $\wv_{2L+1}$ (which is updated at time $2L$) is used. Generally, for any time step $t \in [T]$, the corresponding global model is equal to $\wv_{\psi_{L}(t)}$. While the communication costs are reduced to $32KD/L$, the use of outdated global model might degrade learning accuracy. Thus, the transmission period $L$ should be carefully determined by considering the tradeoff between learning accuracy and communication overhead. On the basis of our analysis, we will give the answer to this question in Section~\ref{subsubsec:pa}.

\vspace{0.1cm}
\noindent{\bf Quantization:} To reduce the communication costs, it is crucial to decrease the number of transmitted bits for conveying the local information $\mv_{k,t}$'s. This can be done from the quantization of the information $\mv_{k,t}$ \cite{reisizadeh2020fedpaq,shlezinger2020uveqfed, oh2022fedvqcs, brinkrolf2021federated}. In this paper, a $q$-bit quantizer $Q_{\rm q}(\cdot)$ is assumed. Note that any scalar or vector quantizer can be applied to the proposed OFL method. For our analysis and experiments, a specific quantizer will be defined.

\vspace{0.1cm}
Incorporating the aforementioned techniques into FedOGD, we present a communication-efficient OFL method, named {\bf O}nline {\bf Fed}erated learning with {\bf I}ntermittent transmission and {\bf Q}uantization (OFedIQ). The proposed method performs with the following two steps (see Fig.~\ref{fig:description}):

\vspace{0.1cm}
\noindent{\bf i) Local update:}
At time step $t \in [T]$, each client $k \in [K]$ optimizes the local model by leveraging the incoming data $(\xv_{k,t},y_{k,t})$ and the up-to-date global model. First, according to the time step $t$, the parameter of a local function is updated via either globally or locally:
\begin{equation}
    \gv_{k,t} = 
    \begin{cases}
    \wv_{t},\; t-1 \in [T]_L\\
    \thetav_{k,t},\; \mbox{else}
    \end{cases} 
\end{equation} with initial values $\wv_1$ and $\thetav_{k,1}$ for $\forall k \in [K]$. Using the incoming data, the local model is updated via OGD:
\begin{equation}\label{eq:3}
    \thetav_{k,t+1} = \gv_{k,t} - \eta \nabla\Lc(f(\xv_{k,t};\gv_{k,t}),y_{k,t}),
\end{equation}
where $\eta>0$ is a learning rate. 

Due to the use of {\em periodic transmission}, at every $L$ time steps (i.e., when $t \in [T]_{L}\subset [T]$), each client $k$ constructs the quantized and weighted local information 
$\mv_{k,t}$ such as
\begin{align}
    \mv_{k,t}&=Q_{\rm q}\left(-\frac{\thetav_{k,t+1}-\gv_{k,t-L+1}}{\eta \times p_k}\right) \label{eq:m}\\
    &=Q_{\rm q}\left( \sum_{\ell=1}^{L}\frac{\nabla\Lc(f(\xv_{k,t-L+\ell};\gv_{k,t-L+\ell}),y_{k,t-L+\ell})}{p_k}\right).\nonumber
\end{align} Moreover, via {\em client subsampling}, every client $k$ can have a chance to participate in the global update with probability $p_k$. That is, at every $L$ time step, each client $k$ can send the local information $\mv_{k,t}$ to the central server with probability $p_k$.

\vspace{0.1cm}
\noindent{\bf ii) Global update:} This step is conducted periodically only when $t \in [T]_{L}$. Let $\Sc_{t}$ denote the index subset of participating (or sampled) clients at time step $t$. Using the received information $\{\mv_{k,t}: k \in \Sc_{t}\}$, at each time step $t \in [T]_{L}$, the central server constructs the global parameter $\wv_{t+1}$ by simply averaging them:
\begin{align}
    \wv_{t+1}&=\wv_{t-L+1} - \frac{\eta}{K}\sum_{k\in\Sc_{t}}\mv_{k,t}\nonumber\\
     &=\wv_{t-L+1}-\frac{\eta}{K}\sum_{k=1}^{K}\mathbbm{1}_{\{k\in \Sc_t\}}\mv_{k,t},\label{eq:2}
\end{align} where $\mathbbm{1}_{\{k\in \Sc_t\}}$ stands for an indicator function having $1$ if the client $k$ participates in the global update at time $t$ (i.e., $k \in \Sc_{t}$).  Then, the central server broadcasts the updated global model $\wv_{t+1}$ to all $K$ clients. We notice that the central server can perform the averaging operation in \eqref{eq:2} without the knowledge of $\Sc_t$, namely, it simply adds up all the received local information. The overall procedures are summarized in {\bf Algorithm 2}.

Recall that the communication costs of the vanilla method (FedOGD) are equal to $32KD$ bits. Whereas, the communication costs of OFedIQ are reduced to $pKD\log{(q)}/L$ bits. On the basis of our regret analysis, the parameters $p$, $L$, and $q$ will be efficiently optimized in 
Section~\ref{subsec:opt}.

\begin{algorithm}
\caption{Proposed Method: OFedIQ}\label{alg:cap}
\begin{algorithmic}[1]
\State {\bf Input:} $K$ clients, sampling rate $0<p_k\leq 1$ for $k \in [K]$, transmission period $L$, quantization parameters (e.g., $(s, b)$ in Definition~\ref{def:quant}), and learning rate $\eta$.
\State {\bf Output:} A sequence of global models $\{\wv_{t+1}:t\in [T]_{L} \}$ (equivalently, $\{\wv_{\psi_L(t)}: t\in [T]\}$).
\State {\bf Initialization:} $\wv_{1}$ and $\thetav_{k,1}$ for $\forall k \in [K]$.
\State {\bf Iteration:} $t=1,2,\ldots,T$
    \Statex$\diamond$ At the client $k\in[K]$
        \Statex\quad$\bullet$ Receive streaming data $(\xv_{k,t},y_{k,t}).$
        \Statex\quad$\bullet$ Update local parameter $\thetav_{k,t+1}$ via \eqref{eq:3}.
        \Statex\quad$\bullet$ Only when $t \in [T]_{L}$ and activated with prob. $p_k$:
        \Statex\quad\quad $-$ Transmit $\mv_{k,t}$ in \eqref{eq:m} to the central server. 
    \Statex$\diamond$ At the central server only when $t \in [T]_{L}$
        \Statex\quad$\bullet$ Receive $\mv_{k,t}$ from some part of clients $k\in\Sc_t$. 
        \Statex\quad$\bullet$ Update the global parameter $\wv_{t+1}$ via \eqref{eq:2}.
        \Statex\quad$\bullet$ Broadcast $\wv_{t+1}$ to the $K$ clients.
\end{algorithmic}


$\Diamond$ OFedIQ with $L=1$ and no quantization is called OFedAvg.


\end{algorithm}


%
%
\section{Regret Analysis}\label{sec:TA}

In this section, we conduct a tighter regret analysis for OFL methods in Section~\ref{sec:methods}. Our analysis is non-trivial extensions of the existing ones such as the centralized OL \cite{bubeck2011introduction,shen2019random,hong2020active} and decentralized OL \cite{hong2020active,chae2021distributed, hong2021communication, shen2021distributed, mitra2021online}. For the first time, our regret bound can reflect the impact of data-heterogeneity and the key communication-reduction techniques such as client subsampling, periodic transmission, and quantization. Contrary to the existing OL analyses \cite{bubeck2011introduction, shen2019random, shen2021distributed, hong2020active, chae2021distributed, hong2021communication}, we do not assume the bounded gradient (a.k.a. $G$-Lipschitz continuous) such as $\|\nabla\Lc(f(\xv_{k,t};\wv),y_{k,t})\|^2\leq G$ for any $\wv, k\in[K], t\in [T]$.
As identified in SGD analysis of FL \cite{khaled2020tighter}, this assumption makes the OL analysis simple but it can ignore the effect of data-heterogeneity. Instead, we introduce a more meaningful quantity (defined by $\sigma_{\rm diff}^2$ in Definition~\ref{def:heterogeneity}), which can capture the degree of data-heterogeneity over time and across decentralized clients. Based on our tighter regret analysis,  we theoretically prove that OFedIQ over $T$ time steps can achieve an optimal subliner regret $\Oc(\sqrt{\alpha T})$) for any dataset, wherein the constant  $\alpha>0$ relies on the degree of data-heterogeneity and the key parameters of the communication-reduction techniques (see Theorem 1 and Theorem 2). As expected, $\alpha$ tends to be bigger as data-heterogeneity (i.e., $\sigma_{\rm diff}^2$) grows. More importantly, our regret bound enables to optimize the key parameters of OFedIQ such as sampling rate, transmission period, and quantization levels (see Section~\ref{subsec:opt}).

Before stating the main results, we provide some useful notations and definitions.  Let $\wv_{\star}$ be the parameter of the best function in hindsight, i.e.,
\begin{equation}\label{eq:w_opt}
    \wv_{\star} \eqdef \argmin_{\wv\in\RR^D}\sum_{t=1}^{T}\sum_{k=1}^{K} \Lc(f(\xv_{k,t};\wv), y_{k,t}).
\end{equation} Recall that $\{\wv_{t+1}: t \in [T]_{L}\}$ is the sequence of outputs provided by Algorithm 2 and accordingly, from the notation in \eqref{eq:notation2}, $\wv_{\psi_L(t)}$ is the common global model for $t \in [T]$. For ease of exposition, let $\EE[\cdot|\wv_{(\lfloor (t-1)/L \rfloor-1)\cdot L +1}]\eqdef\EE_{t}[\cdot]$ be the conditional expectation at time step $t$ given the latest global model. Also, we let
\begin{align*}
    p_{\rm min}&\eqdef\min\{p_1,p_2,...,p_K\}\mbox{ and }p_{\rm sum}\eqdef\sum_{k=1}^{K}p_k,
\end{align*} where $p_{\rm min}$ can capture the effect of the slowest client (e.g., straggler), and $p_{\rm sum}K$ denotes the number of clients (in average) that participates in a global model update at every time step. Extending the idea in \cite{khaled2020tighter} into OFL setting having no statistical assumption on data, we have:
\begin{definition}\label{def:heterogeneity} Given decentralized streaming data $\{(\xv_{k,t}, y_{k,t}): t\in [T], k \in [K]\}$, we define the measure of variance at the optimum:
\begin{align*}
    \sigma^2_{\rm diff}\eqdef\frac{1}{K}\sum_{k=1}^{K}\left[\frac{1}{T}\sum_{t=1}^{T}\left\|\nabla\Lc(f(\xv_{k,t};\wv_{\star}),y_{k,t})\right\|^2\right],
\end{align*} where $\wv_{\star}$ is defined in \eqref{eq:w_opt}.
\end{definition} This quantity reflects the degree of data-heterogeneity (or environmental dynamics in data acquisitions) in both temporal (i.e., over time) and geographical (i.e., across decentralized clients).

\subsection{Main Results}\label{subsec:main}

For our analysis, we make the following assumptions, which are commonly assumed for the analysis of online convex optimization and online learning.


\vspace{0.1cm}
\noindent{\bf Assumption 1.} For any fixed $(\xv_{k,t},y_{k,t})$, the loss function $\Lc(f(\xv_{k,t};\wv), y_{k,t})$ is convex with respect to $\wv$ and differentiable.

\vspace{0.1cm}
\noindent{\bf Assumption 2.} For any fixed $(\xv_{k,t},y_{k,t})$, the loss function $\Lc(f(\xv_{k,t};\wv), y_{k,t})$ is $\beta$-smooth.
\vspace{0.2cm} 

\noindent We state the main results of this section.


\begin{theorem} Under Assumption 1 and Assumption 2, {\bf OFedAvg} with $\eta < \frac{p_{\rm min}}{2\beta}$ (i.e., $L=1$ and no quantization), achieves the following regret bound:
\begin{align*}
    \mbox{regret}_{T}&= \sum_{t=1}^{T}\sum_{k=1}^{K}\EE_{t}\left[\Lc\left(f(\xv_{k,t};{\wv}_{t}),y_{k,t} \right)\right]\nonumber\\
    &\quad\quad\quad\quad\quad\quad-\sum_{t=1}^{T}\sum_{k=1}^{K}\Lc\left(f(\xv_{k,t};{\wv}_{\star}),y_{k,t} \right)\\
    &\leq \frac{K\|\wv_{\star}\|^2}{2\eta} + \frac{\eta KT\sigma_{\rm diff}^2}{p_{\rm min}}.
\end{align*} Note that {\bf FedOGD} achieves the above regret bound with $p_{\rm min}=1$.
\end{theorem}
\begin{IEEEproof}
The proof is provided in Section~\ref{proof2}.
\end{IEEEproof}

\noindent From Theorem 1 and the regret bound of OGD in \cite{bubeck2011introduction, hong2020active}, we obtain:
\begin{align*}
\Delta_{\rm F-C}&=L_{\rm FedOGD} - L_{\rm OGD}\\
&\leq \frac{\left(1-1/\sqrt{K}\right)\sqrt{2\|\wv_{\star}\|^2 \sigma_{\rm diff}^2}}{\sqrt{T}},
\end{align*} where $L_{\rm FedOGD}$ and $L_{\rm OGD}$ denote the average cumulative losses of FedOGD and OGD, respectively. This gap reveals the performance loss caused by distributed learning under OL framework. As $K$ grows, the gap becomes larger but it is bounded as
\begin{equation}
\Delta_{\rm F-C} \leq\sqrt{\frac{2\|\wv_{\star}\|^2 \sigma_{\rm diff}^2}{T}}.
\end{equation} For a sufficiently large $T$, FedOGD can achieve the performance of the centralized OGD while preserving local data-privacy.

We next investigate if OFedIQ still guarantees the asymptotic optimality while reducing the communication costs of FedOGD. Toward this, we derive the regret bound of OFedIQ under the stochastic quantization below.

\vspace{0.1cm}
\noindent{\bf Assumption 3.} $Q_{\rm q}$ is an unbiased $q$-bit stochastic quantizer with a bounded variance $\sigma^2_{\rm q}>0$, i.e., 
\begin{align*}
    \EE\left[Q_{\rm q}(\uv)-\uv\right]=0 \mbox{ and } \EE\left[\|Q_{\rm q}(\uv)-\uv\|^2\right]\leq \sigma_{\rm q}^2\|\uv\|^2.
\end{align*}

\begin{theorem} Under Assumption 1 - Assumption 3, {\bf OFedIQ} with $\eta < \frac{p_{\rm min}}{4\beta(1+\sigma_{\rm q}^2(p_{\rm sum}-p_{\rm min}+1)/K)}$ achieves the following regret bound:
\begin{align*}
&\mbox{regret}_{T}=\sum_{t=1}^{T}\sum_{k=1}^K\;\EE_{t}\left[\Lc\left(f(\xv_{k,t};\wv_{\psi_L(t)}),y_{k,t}\right)\right]\nonumber\\
&\quad\quad\quad\quad\quad\quad\quad\quad-\sum_{t=1}^{T}\sum_{k=1}^{K}\Lc\left(f(\xv_{k,t};\wv_{\star}),y_{k,t}\right)\\
&\leq\frac{K\|\wv_{\star}\|^2}{2\eta}+\frac{2\eta L  K T \sigma_{\rm diff}^2}{p_{\rm min}}\left[1+\frac{\sigma_{\rm q}^2(p_{\rm sum}-p_{\rm min}+1)}{K}\right]\nonumber\\
&\quad\quad+3 \beta \eta^2 L(L-1) K T \sigma_{\rm diff}^2.\nonumber
\end{align*} 
\end{theorem}
\begin{IEEEproof}
The proof is provided in Section~\ref{proof3}.
\end{IEEEproof} 
\vspace{0.1cm}
Note that when quantization and periodic transmission are not used (i.e., $\sigma_q^2=0$ and $L=1$), the regret bound in Theorem 2 is not exactly matched with that in Theorem 1 although in this case, OFedIQ is reduced to OFedAvg. Specifically, there is the difference in the second term by the factor of 2. In fact, this is due to the mathematical artifact in our bounding technique to deal with quantized gradients. Thus, the regret bound of OFedIQ with no quantization is obtained from Theorem 2 with $\sigma_q^2=0$ and by removing the constant 2 in the second term.

\renewcommand{\arraystretch}{1.2} 
\begin{table*}\label{tb:regret}
{\caption{Comparisons of Performances and Communication Costs\\ for Various OFL Methods When $p_k=p, \forall k$ and $(s,b)$-stochastic quantizer is used}} 
\label{tb:overhead}
\centering
\begin{tabular}{ c|| c ||c } 
\Xhline{2\arrayrulewidth}
{Algorithms} & {\bf Regret Bound} $\left(\sqrt{\alpha \|\wv_{\star}\|^2 K^2 \sigma_{\rm diff}^2 T}\right)$ & {\bf Communication Costs (per time step)}\\
\Xhline{2\arrayrulewidth}
{\bf FedOGD} & $\alpha=2$ & $32KD$ bits   \\ 
\hline
{\bf OFedAvg} & $\alpha=2/p$ & $32pKD$ bits   \\ 
\hline
{\bf OFedIQ$(L,p,s,b)$} & $\alpha=4\left(3p^2(L-1)/8+1+\sqrt{D/s^2b}(p+1/K)\right)/(p/L)$ & $(p/L)K(32b+D(1+\log{(s+1)}))$ bits \\ 
\Xhline{2\arrayrulewidth}
\end{tabular}
\end{table*}
\subsection{Optimization of OFedIQ}\label{subsec:opt}
Although the asymptotic optimality of OFedIQ is proved, it does not ensure the best performances of OFedIQ for real-world OFL tasks. Thus, on the basis of our regret analysis in Section~\ref{subsec:main}, we further optimize the parameters of OFedIQ in Algorithm 2 such as sampling rate, transmission period, and quantization parameters. First we introduce a stochastic quantizer that will be used for our experiments in Section~\ref{sec:Exp}. This is an extension of the widely-used stochastic quantizer in \cite{alistarh2017qsgd, reisizadeh2020fedpaq} by introducing the new parameter $b$ (see Definition~\ref{def:quant} for details), where the proposed quantizer with $b=1$ is reduced to the existing one in \cite{alistarh2017qsgd, reisizadeh2020fedpaq}. With the optimized parameters, it will be shown that the proposed quantizer in Definition~\ref{def:quant} can attain a non-trivial gain compared with the existing one.


\begin{definition}\label{def:quant} {\em $(s,b)$-stochastic quantizer $Q_{s,b}(\cdot)$}: For any positive integer $b$, let $\Ic_1,\Ic_2,...,\Ic_b$ be the partition of the indices of the vector $\uv \in \RR^D$ (i.e., the partition of $[D]$) such that
\begin{equation*}
    \Ic_{\ell} \eqdef \left\{\sum_{\tau=1}^{\ell-1}|\Ic_\tau|+1,....,\sum_{\tau=1}^{\ell}|\Ic_\tau|\right\}.
\end{equation*} Note that in our experiments, the uniform-size partition is considered, i.e., $|\Ic_{\ell}|\approx \frac{D}{b}$ for $\forall \ell \in [b]$.
Given a vector $\uv \in \RR^D$, let $\uv_{\Ic_\ell}$ be the subvector of $\uv$, which is formed by taking the entries of $\uv$ whose indices belong to $\Ic_{\ell}$. For any  $\uv \in \RR^D$, stochastic quantizer $Q_{s,b}:\RR^D \rightarrow \RR^D$ is defined as
\begin{equation}
    \left(Q_{s,b}(\uv)\right)_i=\|\uv_{\Ic_\ell}\| \cdot\mathbf{sign}(\uv_i)\cdot\xi_i(\uv,s),
\end{equation}for $i \in \Ic_\ell$, where $\uv_i$ is the $i$-th component of $\uv$ and $\xi_i(\uv,s)$ is a random variable with
\begin{equation*}
    \xi_i(\uv,s)=\begin{cases}
    m/s,       & \quad \mbox{with prob. } 1-\left(\frac{|\uv_i|}{\|\uv_{\Ic_\ell}\|}s-m\right)\\
    (m+1)/s,  & \quad \text{otherwise}. 
  \end{cases}
\end{equation*}
Here, $m \in [0,s)$ is an integer such that $|\uv_i|/\|\uv_{\Ic_{\ell}}\| \in[m/s,\,(m+1)/s]$ for $\forall i \in \Ic_{\ell}$. The parameters $s$ and $b$ denote the quantization-level for each entry of $\uv$ and the number of partitions to indicate magnitudes, respectively. Then, $Q_{s,b}(\cdot)$ is the $q$-bit stochastic quantizer with 
\begin{equation}
    q=32b + D(1+\log{(s+1)}),
\end{equation} where $32b$ bits are used to represent the $b$ real values $\|\uv_{\Ic_{\ell}}\|, \ell=1,...,b$, precisely, and $D(1+\log{(s+1)})$ bits to indicate the sign and $s$-level quantization for each entry of $\uv$. Note that when $b=1$, this quantizer is reduced to the $s$-level stochastic quantizer introduced in \cite{reisizadeh2020fedpaq}. Also, we can verify that $(s,b)$-stochastic quantizer $Q_{s,b}$ satisfies the Assumption 3 such as
\begin{align*}
    \EE\left[Q_{s,b}(\uv) - \uv\right] &= 0,\\
    \EE[\|Q_{s,b}(\uv) - \uv\|^2]&\stackrel{(a)}{\leq} \sum_{\ell=1}^{b}\min\left\{|\Ic_{\ell}|/s^2,\,\sqrt{|\Ic_{\ell}|}/s\right\} \|\uv_{\Ic_{\ell}}\|^2\\
    &\stackrel{(b)}{=}\min\left\{\lceil D/b \rceil/s^2, \sqrt{\lceil D/b \rceil}/s\right\}\|\uv\|^2,
\end{align*} where (a) is directly from the analysis of $s$-level stochastic quantizer in \cite{reisizadeh2020fedpaq} and (b) is due to the fact that $|\Ic_{\ell}|\leq \lceil D/n \rceil$ (i.e., the uniform-size partition). Then, we have:
\begin{equation*}
    \sigma_{\rm q}^2\eqdef \min\left\{\lceil D/b \rceil/s^2, \sqrt{\lceil D/b \rceil}/s\right\}.
\end{equation*} 
\end{definition} Throughout the paper, OFedIQ using the quantizer in Definition 2 is referred to as OFedIQ$(L, p, s, b)$. To optimize the four parameters, we first specify the regret bounds in Theorem 1 and Theorem 2 using the optimized learning rates.
\vspace{0.1cm}
\begin{corollary} Suppose that $p_k=p, \forall k \in [K]$ and $(s,b)$-stochastic quantizer in Definition~\ref{def:quant} is used. In this setting, the regret bounds of OFedAvg and OFedIQ in Theorem 1 and Theorem 2 can be simplified as follows. To clarify the differences, the regret bounds are expressed as $\mbox{regret}_{T}\leq \sqrt{\alpha \|\wv_{\star}\|^2 K^2  \sigma_{\rm diff}^2 T}$, where the constant $\alpha$ relies on the OFL methods:
\begin{itemize}
\item OFedAvg with $\eta=\sqrt{\frac{p\|\wv_{\star}\|^2}{2\sigma_{\rm diff}^2T}}$:
\begin{equation*}
\alpha = 2/p. 
\end{equation*}
\item OFedIQ with $\eta=\sqrt{\frac{(p/L)\|\wv_{\star}\|^2}{4  \sigma_{\rm diff}^2 (3p^2(L-1)/8+1+C) T }}$:
\begin{equation*}
\alpha=\frac{4L}{p}\left(\frac{3p^2(L-1)}{8}+1+\sqrt{D/bs^2}(p+1/K)\right),
\end{equation*} where $C=1+\sqrt{D/bs^2}(p+1/K)$.
\end{itemize} The corresponding communication costs are provided in Table 1.
\end{corollary}
\begin{IEEEproof}
The proof is immediately done from Theorem 1 and Theorem 2, and using the fact that $\sigma_{\rm q}^2\leq \sqrt{D/bs^2}$. Also, the following inequality is used:
\begin{align*}
    3\beta \eta^2 L (L-1) K T \sigma_{\rm diff}^2 &\stackrel{(a)}{\leq} 
    \left(3p^2(L-1)/8\right)\frac{2\eta L K T \sigma_{\rm diff}^2}{p},
\end{align*} where (a) is due to the fact that $\eta < \frac{p}{4\beta}$.
\end{IEEEproof}

\vspace{0.2cm}
From Corollary 1, we will verify that periodic transmission is unnecessary, namely, $L=1$ is the best choice for OFedIQ (see Section~\ref{subsubsec:pa}). Furthermore, we will optimize the remaining parameters $p$ and $(s,b)$ in Section~\ref{subsubsec:op}.

\subsubsection{Needlessness of periodic transmission}\label{subsubsec:pa}

Consider the two OFedIQs having the same communication costs such as OFedIQ$(L,p,s,b)$ and OFedIQ$(L=1,p'=p/L,s,b)$. Leveraging the regret bounds in Table 1, we can identify that the latter outperforms the former as follows:
\begin{align*}
    &\frac{1+\sqrt{D/s^2b}(p'+1/K)}{p'} = \frac{1+\sqrt{D/s^2b}(p/L+1/K)}{p/L}\nonumber\\
    &\quad\quad\leq\frac{3p^2(L-1)/8+1+\sqrt{D/s^2b}(p+1/K)}{p/L},
\end{align*} where the equality holds with $L=1$. This analysis shows that in OFL, client subsampling is sufficient for intermittent transmission, namely, periodic transmission is unnecessary. Thus, $L=1$ is the best choice for OFedIQ. In the subsequent subsection, we will optimize the remaining parameters $p$ and $(s,b)$ of OFedIQ.


\subsubsection{Parameter Optimization}\label{subsubsec:op}

We first show the necessity of quantization to reduce the communication costs. To this end, we compare the regret bounds of OFedAvg$(p)$ and OFedIQ$(L=1,p',s=\lceil 40p \rceil,b=10^{-2}D)$. Having the same communication costs, the parameter $p'$ should be chosen as the function of $p$ such as
\begin{align}
    p'&=\frac{32pD}{32b+D(1+\log(1+s))}\stackrel{(a)}{=}\frac{32p}{1.32+\log(1+s)},\label{eq:pp1}
\end{align} where (a) is due to the fact that $b=10^{-2}D$. Also, from Table 1, the regret bound of OFedIQ$(L=1,p',s=\lceil 40p \rceil,b=10^{-2}D)$ is obtained as follows:
\begin{align}
\alpha &=\frac{2}{p'}\left[1+\frac{10}{s}\left(p' + \frac{1}{K}\right)\right]\nonumber\\
    &\stackrel{(a)}{\leq} \frac{2}{p'} + \frac{40}{s}  \nonumber\\
    &\stackrel{(b)}{\leq}\frac{1}{p}\left[\frac{1.32}{16}+\frac{1}{16}\log(20p+1)\right] + \frac{1}{p}\nonumber\\
    &\stackrel{(c)}{\leq} \frac{1}{p}\left[\frac{1.32}{16}+\frac{\log(20+1)}{16}+1\right]<\frac{1.3570}{p},\label{eq:regret-OFedAQ}
\end{align} where (a) is due to the reasonable assumption of $K \gg 1/p'$, (b) is from \eqref{eq:pp1} and due to the fact that $s=\lceil 40p \rceil \leq 40p$, and (c) is from $p<1$. This shows that the regret bound of OFedIQ with the above parameters is lower than that of OFedAvg, while having the same communication costs. Although the parameters of our quantizer is not optimized, OFedIQ with quantization can outperform OFedAvg. Thus, we can conclude that quantization is indeed necessary.


Next, given target communication costs (say, $32\gamma KD$), we optimize the parameters of OFedIQ$(L=1,p,s,b=\rho D)$, where $0<\gamma<1$ controls the reduction of communication costs compared with FedOGD (the baseline method) and $b=\rho D$ is chosen from the analysis in \eqref{eq:regret-OFedAQ}. To yield the given communication costs $\gamma$, the parameters $p$ and $(s,b=\rho D)$ should satisfy the equality:
\begin{equation}\label{eq:gamma}
    \gamma = p(32\rho+(1+\log(s+1))/32.
\end{equation} Plugging \eqref{eq:gamma} into the regret bound of OFedIQ in Table 1, the parameter optimization problem can be formulated as
\begin{align}
    \min_{\rho, s}\; 2\rho/\gamma + \log(s+1)/16 \gamma + \sqrt{4/s^2\rho}.\label{eq:param_opt}
\end{align} First of all, the objective function in \eqref{eq:param_opt} is minimized if $\rho$ is chosen such that
\begin{equation}
    2\rho/\gamma = \sqrt{4/s^2\rho},
\end{equation} which yields
\begin{equation}\label{eq:gamma1}
    \rho = \left(\gamma/s\right)^{\frac{2}{3}}.
\end{equation} Using this $\rho$, the optimization problem in \eqref{eq:param_opt} can be rewritten as
\begin{equation}
    \min_{s}\; \log(s+1)/16 + 4\left(\gamma/s\right)^{2/3}.\label{eq:gamma2}
\end{equation} This problem is easily solved via the simple search over $s \in \ZZ^{+}$. To sum up, given the communication costs $0<\gamma\leq 1$, the algorithm to optimize the parameters $p$ and $(s,b=\rho D)$ are described in {\bf Algorithm~\ref{alg:parameter-opt}}.
In the example of $\gamma=0.1$ and $D=34,826$, via Algorithm~\ref{alg:parameter-opt}, the parameters of OFedIQ$(L,p,s,b)$ are optimized as follows:
\begin{align*}
    &s=17, \rho=0.0326, b=\lfloor 0.0326\times D\rfloor=1134,\\
    &p=0.5151, \mbox{and } L=1.
\end{align*} We observe that the regret bound of OFedIQ with the optimized parameters is equal to $4.536$, which is much lower than that of OFedAvg (e.g., $20$). This comparison can strengthen our argument that quantization is indeed necessary.
\begin{algorithm}
\caption{Proposed Parameter Optimization}\label{alg:parameter-opt}
\begin{algorithmic}[1]
\State {\bf Input:} Communication costs $0<\gamma\leq 1$ and model size $D$.
\State {\bf Output:} The parameters $p$ and $(s,b)$.
\State {\bf Procedure:}
    \Statex$\bullet$ Find $s \in \ZZ^{+}$ by taking the solution of \eqref{eq:gamma2}.
    \Statex$\bullet$ Determine $b=\lfloor (\gamma/s)^{2/3} D \rfloor$.
    \Statex$\bullet$ Determine $p=32\gamma/(1+32 (\gamma/s)^{2/3}+\log(s+1))$.
\end{algorithmic}
\end{algorithm}

\subsection{Proof of Main Theorems}

We prove the regret bounds in Theorem 1 and Theorem 2. To simplify the expressions in the proofs, we let
\begin{equation}
    \nabla_{k,t}(\wv)\eqdef\nabla\Lc\left(f(\xv_{k,t};{\wv}),y_{k,t} \right),
\end{equation}for any vector $\wv\in\RR^D$. We first provide the supporting lemmas (Lemma 1 and Lemma 2) for the proofs.
\begin{lemma}\label{lem1}\normalfont \
    Letting $\vv_i,\vv_j\in\mathbb{R}^d$ and for any $a>0$, the following inequality holds:
    \begin{align*}
    \|\vv_i+\vv_j\|^2\leq(1+a)\|\vv_i\|^2+\left(1+1/a\right)\|\vv_j\|^2.
    \end{align*}
\end{lemma}
\begin{IEEEproof} 
        The proof is directly from the equality: $\|\vv_i+\vv_j\|^2=(1+a)\|\vv_i\|^2+\left(1+1/a\right)\|\vv_j\|^2
            -\left\|\sqrt{a}\vv_i-\vv_j/\sqrt{a}\right\|^2$.
\end{IEEEproof} 
\begin{lemma}\label{lem2} The following inequality holds:
\begin{align*}
    &\frac{1}{K}\sum_{k=1}^{K}\|\nabla_{k,t}(\wv_t)\|^2\nonumber\\
    &\leq \frac{4\beta}{K}\sum_{k=1}^{K}[\Lc\left(f(\xv_{k,t};{\wv}_{\star}),y_{k,t} \right) -\Lc\left(f(\xv_{k,t};{\wv}_{t}),y_{k,t} \right)]\nonumber\\
    &+\frac{4\beta}{K}\sum_{k=1}^{K} \nabla_{k,t}(\wv_{t})^{\trasp}(\wv_{t}-\wv_{\star})+\frac{2}{K}\sum_{k=1}^{K}\|\nabla_{k,t}(\wv_{\star})\|^2.
\end{align*}
\end{lemma}
\begin{IEEEproof} From Lemma~\ref{lem1} with $a=1$, we have:
\begin{align}
    \frac{1}{K}\sum_{k=1}^{K}\|\nabla_{k,t}(\wv_t)\|^2&\leq \frac{2}{K}\sum_{k=1}^{K}\left\|\nabla_{k,t}(\wv_{t})-\nabla_{k,t}(\wv_{\star})\right\|^2\nonumber\\
    &+\frac{2}{K}\sum_{k=1}^{K}\left\|\nabla_{k,t}(\wv_{\star})\right\|^2.\label{eq:lem2-1}
\end{align} Also, from Assumption 2 and \cite[Theorem 2.15]{nesterov2018lectures}, we obtain the inequality below:
\begin{align}
    &\left\|\nabla_{k,t}(\wv_{t})-\nabla_{k,t}(\wv_{\star})\right\|^2\nonumber\\
    &\quad\leq 2\beta\left[ \Lc\left(f(\xv_{k,t};{\wv}_{\star}),y_{k,t} \right) -\Lc\left(f(\xv_{k,t};{\wv}_{t}),y_{k,t} \right)\right] \nonumber\\
    &\quad\quad + 2\beta\nabla_{k,t}(\wv_{t})^{\trasp}(\wv_{t}-\wv_{\star})]. \label{eq:lem2-2}
\end{align} Plugging \eqref{eq:lem2-2} into \eqref{eq:lem2-1}, the proof of Lemma 2 is completed.
\end{IEEEproof}

\subsubsection{Proof of Theorem 1}\label{proof2}
Note that the supporting lemma (Lemma 2) for this proof is provided in the last part of this subsection. From the global update rule of OFedAvg in \eqref{eq:2}, we have:
\begin{align}
    \wv_{t+1} 
     &=\wv_{t} - \frac{\eta}{K}\sum_{k=1}^{K}\mathbbm{1}_{\{k\in\Sc_t\}}\frac{\nabla_{k,t}(\wv_{t})}{p_k},\label{eq:OFedAvgUpdate}
\end{align} where $\mathbbm{1}_{\{k\in\Sc_t\}}$ is an indicator function having $1$ if the client $k$ is active at time $t$. From \eqref{eq:OFedAvgUpdate}, taking the conditional expectation given the latest global parameter $\wv_{t}$ (in short, $\EE_{t+1}[\cdot|\wv_{t}]$), we obtain the key inequality: 
\begin{align}
    &\EE_{t+1}\left[\|\wv_{t+1}-\wv_{\star}\|^2\right]\nonumber\\
    &=\|\wv_{t}-\wv_{\star}\|^2+\EE_{t+1}\left[\left\| \frac{\eta}{K}\sum_{k =1}^{K}\mathbbm{1}_{\{k\in\Sc_t\}}\frac{\nabla_{k,t}(\wv_{t})}{p_k}\right\|^2 \right]\nonumber\\
   &-\frac{2\eta}{K}\EE_{t+1}\left[\sum_{k =1}^{K}\mathbbm{1}_{\{k\in\Sc_t\}}\left(\frac{\nabla_{k,t}(\wv_t)}{p_k}\right)^{\trasp}(\wv_{t}-\wv_{\star}) \right].\label{eq:th2-0}
\end{align} 
From the probabilistic client subsampling, we can get:
\begin{align}
    &\EE_{t+1}\left[\sum_{k =1}^{K}\mathbbm{1}_{\{k\in\Sc_t\}}\left(\frac{\nabla_{k,t}(\wv_t)}{p_k}\right)^{\trasp}(\wv_{t}-\wv_{\star}) \right]\nonumber\\
    &\quad\quad\quad\quad\quad=\sum_{k=1}^{K}\nabla_{k,t}(\wv_t)^{\trasp}(\wv_{t}-\wv_{\star}),\label{eq:th2-1}
\end{align} since $\EE_{t+1}[\sum_{k =1}^K \mathbbm{1}_{\{k\in\Sc_t\}}\nabla_{k,t}(\wv_t)/p_k]=\sum_{k=1}^{K}\nabla_{k,t}(\wv_t)$, where the equality is from $\EE[\mathbbm{1}_{\{k\in\Sc_t\}}]=p_k$.
Also, we obtain the following upper-bound:
\begin{align}
        &\EE_{t+1}\left[\left\| \frac{\eta}{K}\sum_{k=1}^{K}\mathbbm{1}_{\{k\in\Sc_t\}}\frac{\nabla_{k,t}(\wv_{t})}{p_k}\right\|^2 \right]\nonumber\\
        &\stackrel{(a)}{\leq} \frac{\eta^2}{K}\sum_{k=1}^{K}\EE_{t+1}\left[\left\|\mathbbm{1}_{\{k\in\Sc_t\}}\frac{\nabla_{k,t}(\wv_t)}{p_k} \right\|^2\right]\nonumber\\
        &\stackrel{(b)}{=}\frac{\eta^2}{K}\sum_{k=1}^{K}\frac{\|\nabla_{k,t}(\wv_t)\|^2}{p_k}\nonumber\\
        &\stackrel{(c)}{\leq}\frac{\eta^2}{Kp_{\rm min}}\sum_{k=1}^{K}\|\nabla_{k,t}(\wv_t)\|^2\label{eq:th2-2}\\
        &\stackrel{(d)}{\leq} \frac{4\eta^2\beta}{K p_{\rm min}}\sum_{k=1}^{K}[\Lc\left(f(\xv_{k,t};{\wv}_{\star}),y_{k,t} \right) -\Lc\left(f(\xv_{k,t};{\wv}_{t}),y_{k,t} \right)]\nonumber\\
        &\quad\quad+\frac{4\eta^2 \beta }{K p_{\rm min}}\sum_{k=1}^{K} \nabla_{k,t}(\wv_{t})^{\trasp}(\wv_{t}-\wv_{\star})\nonumber\\
        &\quad\quad+\frac{2\eta^2}{p_{\rm min}}\left[\frac{1}{K}\sum_{k=1}^{K}\|\nabla_{k,t}(\wv_{\star})\|^2\right],\label{eq:th2-3}
\end{align} where (a) is from Jensen's inequality,  (b) is due to the fact that $\EE[(\mathbbm{1}_{\{k\in\Sc_t\}})^2]=p_k$, (c) is because $p_{\rm min}\leq p_k$ for $\forall K \in [K]$, and (d) is from Lemma~\ref{lem2}.
Plugging \eqref{eq:th2-1} and \eqref{eq:th2-3} into \eqref{eq:th2-0}, we obtain the following upper-bound:
\begin{align}
    &\sum_{k=1}^{K} \nabla_{k,t}(\wv_{t})^{\trasp}(\wv_{t}-\wv_{\star})\stackrel{(a)}{\leq}\nonumber\\
    &\frac{K}{2\eta(1-2\eta\beta/p_{\rm min} )}\left(\|\wv_t - \wv_{\star}\|^2 - \EE_{t+1}\left[\|\wv_{t+1}-\wv_{\star}\|^2\right]\right)\nonumber\\
    &\quad+\frac{2\eta\beta/p_{\rm min}}{(1-2\eta \beta/p_{\rm min} )}\sum_{k=1}^{K}\Big[\Lc\left(f(\xv_{k,t};{\wv}_{\star}),y_{k,t} \right)\nonumber\\
&\quad\quad\quad\quad\quad\quad\quad\quad\quad\quad\quad\quad\quad\quad\quad-\Lc\left(f(\xv_{k,t};{\wv}_{t}),y_{k,t} \right)\Big]\nonumber\\
    &\quad+\frac{\eta/p_{\rm min}}{(1-2\eta \beta/p_{\rm min} )}\sum_{k=1}^{K}\|\nabla_{k,t}(\wv_{\star})\|^2,\label{eq:th1-4}
\end{align} where (a) is due to the condition of $\eta < \frac{p_{\rm min}}{2\beta}$. Also, from Assumption 1 (i.e., the convexity of our loss function), we obtain the lower-bound:
\begin{align}
&\sum_{k=1}^{K}\nabla_{k,t}(\wv_t)^{\trasp}(\wv_t-\wv_{\star})\nonumber\\
&\geq\sum_{k=1}^{K}[\Lc\left(f(\xv_{k,t};{\wv}_{t}),y_{k,t} \right) -\Lc\left(f(\xv_{k,t};{\wv}_{\star}),y_{k,t} \right)].\label{eq:th1-5}
\end{align} Combining the upper-bound in \eqref{eq:th1-4} and lower-bound in \eqref{eq:th1-5}, and taking the conditional expectation $\EE_{t}[\cdot]$ in the both sides, we can get:
\begin{align*}
&\sum_{k=1}^{K}\EE_{t}\left[\Lc\left(f(\xv_{k,t};{\wv}_{t}),y_{k,t} \right)\right] -\Lc\left(f(\xv_{k,t};{\wv}_{\star}),y_{k,t} \right)\\
    &\leq\frac{K}{2\eta}\left(\EE_{t}\left[\|\wv_t - \wv_{\star}\|^2\right] - \EE_{t+1}\left[\|\wv_{t+1}-\wv_{\star}\|^2\right]\right)\\
    &\quad+\frac{\eta}{p_{\rm min}} \sum_{k=1}^{K}\|\nabla_{k,t}(\wv_{\star})\|^2,
\end{align*} where we used the fact that $\EE\left[\|\wv_{t+1}-\wv_{\star}\|^2|\wv_t, \wv_{t-1}\right] = \EE\left[\|\wv_{t+1}-\wv_{\star}\|^2|\wv_t\right]$ as $\wv_{t+1}$ and $\wv_{t-1}$ are conditionally independent given the knowledge of $\wv_{t}$. Summing over $t=1,2,...,T$ and using the telescoping sum, we obtain the following regret bound:
\begin{align*}
&\sum_{t=1}^{T}\sum_{k=1}^{K} \EE_{t}\left[\Lc\left(f(\xv_{k,t};{\wv}_{t}),y_{k,t} \right)\right] -\Lc\left(f(\xv_{k,t};{\wv}_{\star}),y_{k,t} \right)\\
&\quad{\leq} \frac{K\|\wv_{\star}\|^2}{2\eta} + \frac{\eta}{p_{\rm min}} \sum_{t=1}^{T}\sum_{k=1}^{K}\|\nabla_{k,t}(\wv_{\star})\|^2\nonumber\\
&\quad\stackrel{(a)}{=}\frac{K\|\wv_{\star}\|^2}{2\eta} +\frac{\eta KT\sigma_{\rm diff}^2}{p_{\rm min}},
\end{align*} where (a) is from Definition 1. This completes the proof of Theorem 1.

\vspace{0.2cm}
\subsubsection{Proof of Theorem 2}\label{proof3}
To clarify the proof, we first prove Theorem 2 in the special case of $L=1$ and then extend it to the general case of $L>1$.

\vspace{0.2cm}
{\bf (a) $L=1$.} Given a quantizer $Q(\cdot)$, a quantized gradient is denoted as
\begin{equation*}
    \hat{\nabla}_{k,t}(\wv_{t})=Q\left(\nabla_{k,t}(\wv_t)/p_k\right).
\end{equation*} From the global update rule of OFedIQ in \eqref{eq:2}, we have:
\begin{align}
    \wv_{t+1} 
    =\wv_{t} - \frac{\eta}{K}\sum_{k=1}^{K}\mathbbm{1}_{\{k\; {\rm active}\}}\hat{\nabla}_{k,t}(\wv_{t}),\label{eq:th3-00}
\end{align} where $\mathbbm{1}_{\{k\in\Sc\}}$ is an indicator function having $1$ if the client $k$ is active at time $t$. 
As derived in \eqref{eq:th2-0}, from \eqref{eq:th3-00}
we derive the key inequality:
\begin{align}
    &\EE_{t+1}\left[\|\wv_{t+1}-\wv_{\star}\|^2\right]\nonumber\\
    &=\|\wv_{t}-\wv_{\star}\|^2+\EE_{t+1}\left[\left\| \frac{\eta}{K}\sum_{k \in \Sc_t}\hat{\nabla}_{k,t}(\wv_{t})\right\|^2 \right]\nonumber\\
    &\quad -\frac{2\eta}{K}\sum_{k=1}^{K}{\nabla}_{k,t}(\wv_t)^{\trasp}(\wv_{t}-\wv_{\star}),\label{eq:th3-0}
\end{align} where we used the fact that
\begin{align}
    &\EE_{t+1}\left[\sum_{k \in \Sc_{t}}\hat{\nabla}_{k,t}(\wv_{t})\right]\stackrel{(a)}{=}\EE\left[ \EE_{t+1}\left[\sum_{k \in \Sc_t}\frac{\nabla_{k,t}(\wv_{t})}{p_k}\middle|\Sc_t\right]\right]\nonumber\\ &\quad\quad=\EE\left[\sum_{k=1}^{K}\mathbbm{1}_{\{k\in\Sc_t\}} \frac{\nabla_{k,t}(\wv_t)}{p_k} \right]\stackrel{(b)}{=}\sum_{k=1}^{K}\nabla_{k,t}(\wv_t),
\end{align} and where (a) is due to the law of iterated expectation $\EE_{t+1}[\hat{\nabla}_{k,t}(\wv_t)]=\nabla_{k,t}(\wv_t)/p_k$, and (b) follows the fact that $p_k = \EE[\mathbbm{1}_{\{k\in\Sc_t\}}]$. Also, we obtain the following inequality:
\begin{align}
      &\EE_{t+1}\left[\left\| \frac{\eta}{K}\sum_{k \in \Sc_t}\hat{\nabla}_{k,t}(\wv_{t})\right\|^2 \right]\nonumber\\
      &\stackrel{(a)}{\leq}2\eta^2\EE_{t+1}\left[\left\|\frac{1}{K}\sum_{k \in \Sc_t }\left(\hat{\nabla}_{k,t}(\wv_{t}) - \frac{\nabla_{k,t}(\wv_t)}{p_k}\right)\right\|^2 \right]\nonumber\\
      &\quad+2\EE_{t+1}\left[\left\| \frac{\eta}{K}\sum_{k\in\Sc_t}\frac{\nabla_{k,t}(\wv_t)}{p_k}\right\|^2 \right]\nonumber\\
      &\stackrel{(b)}{\leq} \frac{2\eta^2}{p_{\rm min}}\left[\frac{\sigma_{\rm q}^2(p_{\rm sum}-p_{\rm min}+1)}{K}+1\right]\frac{1}{K}\sum_{k=1}^{K}\left\|\nabla_{k,t}(\wv_t)\right\|^2,\label{eq:th3-3}
\end{align} where (a) is from Lemma~\ref{lem1} with $a=1$ and (b) is from \eqref{eq:th2-2} and  the following fact:
\begin{align}
    &\EE_{t+1}\left[\left\|\frac{1}{K}\sum_{k \in \Sc_t }\left(\hat{\nabla}_{k,t}(\wv_{t}) - \frac{\nabla_{k,t}(\wv_t)}{p_k}\right)\right\|^2 \right]\nonumber\\
    &=\EE\left[\frac{1}{K^2}\EE_{t+1}\left[\left\|\sum_{k \in \Sc_t }\left(\hat{\nabla}_{k,t}(\wv_{t}) - \frac{\nabla_{k,t}(\wv_t)}{p_k}\right)\right\|^2 \middle|\Sc_t\right] \right]\nonumber\\
    &\stackrel{(a)}{\leq}\EE\left[\frac{|\Sc_t|}{K^2}\EE_{t+1}\left[\sum_{k \in \Sc_t }\left\|\left(\hat{\nabla}_{k,t}(\wv_{t}) - \frac{\nabla_{k,t}(\wv_t)}{p_k}\right)\right\|^2 \middle|\Sc_t\right]\right]\nonumber\\
    &\stackrel{(b)}{\leq}\frac{\sigma_{\rm q}^2}{K^2} \EE\left[{|\Sc_t|}\EE_{t+1}\left[\sum_{k \in \Sc_t }\left\|\frac{\nabla_{k,t}(\wv_t)}{p_k} \right\|^2 \middle|\Sc_t\right]\right]\nonumber\\
    &=\frac{\sigma_{\rm q}^2}{K^2}\EE\left[\left(\sum_{k=1}^{K}\mathbbm{1}_{\{k\in\Sc_t\}}\right)\left(\sum_{k=1}^{K}\mathbbm{1}_{\{k\in\Sc_t\}}\left\|\frac{\nabla_{k,t}(\wv_t)}{p_k} \right\|^2\right)\right]\nonumber\\
    &\stackrel{(c)}{=}\frac{\sigma_{\rm q}^2}{K^2}\sum_{k=1}^{K}\left(\frac{p_{\rm sum}-p_k +1}{p_{k}}\right)\left\|\nabla_{k,t}(\wv_t) \right\|^2\nonumber\\
    &\leq\frac{\sigma_{\rm q}^2}{K^2}\left(\frac{p_{\rm sum}- p_{\rm min}+1}{p_{\rm min}}\right)\sum_{k=1}^{K}\left\|\nabla_{k,t}(\wv_t) \right\|^2,\label{eq:th-quant}
\end{align}where (a) is from the Jensen's inequality, (b) is from Assumption 4 (i.e., the variance of our quantizer is upper-bounded as $\sigma_{\rm q}^2$), and (c) follows our probabilistic client subsampling. As in the proof of Theorem 1, from \eqref{eq:th3-0} and \eqref{eq:th3-3}, and Lemma 2, we derive the upper-bound:
\begin{align}
    &\sum_{k=1}^{K} \nabla_{k,t}(\wv_{t})^{\trasp}(\wv_{t}-\wv_{\star})\stackrel{(a)}{\leq}\nonumber\\
    &\;\;\frac{K}{2\eta(1-4\eta\beta R/p_{\rm min} )}\left(\|\wv_t - \wv_{\star}\|^2 - \EE_{t+1}\left[\|\wv_{t+1}-\wv_{\star}\|^2\right]\right)\nonumber
    \end{align}
    \begin{align}
    &\quad+\frac{4\eta\beta R/p_{\rm min}}{(1-4\eta \beta R/p_{\rm min} )}\sum_{k=1}^{K}\Big[\Lc\left(f(\xv_{k,t};{\wv}_{\star}),y_{k,t} \right)\nonumber\\
&\quad\quad\quad\quad\quad\quad\quad\quad\quad\quad\quad\quad\quad\quad\quad-\Lc\left(f(\xv_{k,t};{\wv}_{t}),y_{k,t} \right)\Big]\nonumber\\
    &\quad+\frac{2\eta R/p_{\rm min}}{(1-4\eta \beta R/p_{\rm min} )}\sum_{k=1}^{K}\|\nabla_{k,t}(\wv_{\star})\|^2,\label{eq:up0}
\end{align} where for simplicity, we let $R\eqdef(\sigma_{\rm q}^2(p_{\rm sum}-p_{\rm min}+1))/K+1$ and (a) is due to the fact that $\eta<\frac{p_{\rm min}}{4\beta R}$. From the lower-bound in \eqref{eq:th1-5} and the upper-bound in \eqref{eq:up0}, and 
taking the conditional expectation $\EE_{t}[\cdot]$ in both sides, we can get: 
\begin{align*}
&\sum_{k=1}^{K}\EE_{t}\left[\Lc\left(f(\xv_{k,t};{\wv}_{t}),y_{k,t} \right)\right] -\Lc\left(f(\xv_{k,t};{\wv}_{\star}),y_{k,t} \right)\\
    &\leq\frac{K}{2\eta}\left(\EE_{t}\left[\|\wv_t - \wv_{\star}\|^2\right] - \EE_{t+1}\left[\|\wv_{t+1}-\wv_{\star}\|^2\right]\right)\\
    &\quad+\frac{2\eta}{p_{\rm min}} \left[\frac{\sigma_{\rm q}^2\left(p_{\rm sum}  - p_{\rm min}+1\right)}{K}+1\right] \sum_{k=1}^{K}\|\nabla_{k,t}(\wv_{\star})\|^2,
\end{align*} where we used the fact that $\EE\left[\|\wv_{t+1}-\wv_{\star}\|^2|\wv_t, \wv_{t-1}\right] = \EE\left[\|\wv_{t+1}-\wv_{\star}\|^2|\wv_t\right]$ since $\wv_{t+1}$ and $\wv_{t-1}$ are conditionally independent given the knowledge of $\wv_{t}$. Summing over $t=1,2,...,T$ and by the telescoping sum, we obtain the regret bound:
\begin{align*}
&\sum_{t=1}^{T}\sum_{k=1}^{K}\EE_{t}\left[\Lc\left(f(\xv_{k,t};{\wv}_{t}),y_{k,t} \right)\right] -\Lc\left(f(\xv_{k,t};{\wv}_{\star}),y_{k,t} \right)\\
    &\leq \frac{K\|\wv_{\star}\|^2}{2\eta} + \frac{ 2\eta KT\sigma_{\rm diff}^2}{p_{\rm min}} \left[\frac{\sigma_{\rm q}^2\left(p_{\rm sum} - p_{\rm min}+1\right)}{K}+1\right].
\end{align*} This completes the proof of Theorem 2 when $L=1$.


\vspace{0.2cm}
{\bf (b) $L>1$.} We extend the above proof for the general case of $L>1$.  The supporting lemmas (Lemma 3 and Lemma 4) are provided in the last part of this section. To concise the expressions and explain the key ides of our proof clearly, it is assumed that $T$ is a multiple of $L$. In a general case, we need to take some exceptions for the few time indices $\lfloor T/L \rfloor\cdot L +1,\lfloor T/L \rfloor\cdot L +2,...,T$. Clearly, the number of these extra-terms is less than $L$. Thus, when $T$ is sufficiently large, the impact of exceptional terms can be negligible. Supporting lemma (Lemma~\ref{lem3}) for this proof are provided in the last part of this section. 
For simplicity, we let
\begin{equation}\label{eq:df1}
   \nabla_{k,t}^{L}\eqdef \sum_{\ell=1}^L\frac{\nabla_{k,t-L+\ell}}{p_k} \mbox{ and } \hat{\nabla}_{k,t}^{L}\eqdef Q\left(\nabla_{k,t}^{L}\right).
\end{equation}

We first focus on time $t \in [T]_L$ where recall that $[T]_{L} \eqdef \{t\in[T]: t \mod L = 0\}$. At this time, every client is aware of the up-to-date global parameter $\wv_{t-L+1}$ and aims at learning an updated global parameter $\wv_{t+1}$. 
As derived in \eqref{eq:th3-0}, from the global update rule in \eqref{eq:2}, we obtain the key inequality:
\begin{align}
    &\EE_{t+1}\left[\|\wv_{t+1}-\wv_{\star}\|^2\right]\nonumber\\
    &=\left\|\wv_{t-L+1}-\wv_{\star}\right\|^2+\EE_{t+1}\left[\left\|\frac{\eta}{K}\sum_{k\in\Sc_t}\hat{\nabla}_{k,t}^L\right\|^2\right]\nonumber\\
     &\quad -\frac{2\eta}{K}\sum_{k=1}^K\sum_{\ell=1}^{L}\left(\nabla_{k,t-L+\ell}\right)^{\intercal}(\wv_{t-L+1}-\wv_{\star}),\label{eq:pf1_1}
\end{align} where we used the fact that
\begin{align*}
&\EE\left[\sum_{k\in\Sc_t}\hat{\nabla}_{k,t}^{L}\right]=\EE\left[\EE\left[\sum_{k\in\Sc_t}\hat{\nabla}_{k,t}^{L}\middle|\Sc_t\right]\right]\\
    &=\EE\left[\sum_{k=1}^K\sum_{\ell=1}^L\mathbbm{1}_{\{k\in\Sc_t\}}\frac{\nabla_{k,t-L+\ell}}{p_k}\right]\\
    &=\sum_{k=1}^K\sum_{\ell=1}^L\PP(k\in\Sc_t)\frac{\nabla_{k,t-L+\ell}}{p_k}=\sum_{k=1}^K\sum_{\ell=1}^L\nabla_{k,t-L+\ell}.
\end{align*} Note that the randomness in the above is caused by a stochastic quantization and the random selection of active clients. From Jensen's inequality and  \eqref{eq:th3-3}, we obtain the upper-bound:
\begin{align}
    &\mathbb{E}_{t+1}\left[\left\Vert\frac{\eta}{K}\sum_{k\in \Sc_t}\hat{\nabla}_{k,t}^{L}\right\Vert^2\right]\leq \label{eq:pf1_3}\\
       &\frac{2\eta^2 L }{p_{\rm min}K}\left[1+\frac{\sigma_{\rm q}^2(p_{\rm sum}-p_{\rm min} + 1)}{K} \right]\sum_{k=1}^{K}\sum_{\ell=1}^{L}\|\nabla_{k,t-L+\ell}\|^2.\nonumber
\end{align} 
From \eqref{eq:pf1_1}, \eqref{eq:pf1_3}, and Lemma 2,  we obtain the upper-bound:
\begin{align}
    &\sum_{k=1}^{K} \nabla_{k,t}(\wv_{t})^{\trasp}(\wv_{t}-\wv_{\star})\stackrel{(a)}{\leq}\nonumber\\
    &\frac{K}{2\eta(1-4\eta\beta L R/p_{\rm min} )}\Big(\|\wv_{t-L+1} - \wv_{\star}\|^2 \nonumber\\
    &\quad\quad\quad\quad\quad\quad\quad\quad\quad\quad\quad- \EE_{t+1}\left[\|\wv_{t+1}-\wv_{\star}\|^2\right]\Big)\nonumber\\
    &+\frac{4\eta\beta L R/p_{\rm min}}{(1-4\eta \beta L R/p_{\rm min} )}\sum_{k=1}^{K}\sum_{\ell=1}^{L}\Big[\Lc\left(f(\xv_{k,t-L+\ell};{\wv}_{\star}),y_{k,t-L+\ell} \right)\nonumber\\
&\quad\quad\quad\quad\quad\quad\quad\quad\quad-\Lc\left(f(\xv_{k,t-L+\ell};{\wv}_{t-L+1}),y_{k,t-L+\ell} \right)\Big]\nonumber\\
    &+\frac{2\eta L R/p_{\rm min}}{(1-4\eta \beta L R/p_{\rm min} )}\sum_{k=1}^{K}\sum_{\ell=1}^{L}\|\nabla_{k,t-L+\ell}(\wv_{\star})\|^2,\label{eq:up}
\end{align} where for simplicity, we let $R=(\sigma_{\rm q}^2(p_{\rm sum}-p_{\rm min}+1))/K+1$. 
From \cite[Theorem 2.15]{nesterov2018lectures} and Assumption 2, we have:
\begin{align}
    &(\nabla_{k,t-L+\ell})^{\trasp}(\wv_{t-L+1}-\wv_{\star}) \nonumber\\
    &\quad\geq\Lc\left(f(\xv_{k,t-L+\ell};\wv_{t-L+1}),y_{k,t-L+\ell}\right)\nonumber\\
    &\quad\quad\quad\quad\quad-\Lc\left(f(\xv_{k,t-L+\ell};\wv_{\star}),y_{k,t-L+\ell}\right)\nonumber\\
    &\quad\quad-\frac{\beta}{2}\|\wv_{t-L+1}-\gv_{k,t-L+\ell}\|^2.\label{eq:pf1_2}
\end{align}
From \eqref{eq:pf1_2} and Lemma 3, we next obtain the following lower-bound:
\begin{align}
&\sum_{k=1}^{K}\sum_{\ell=1}^{L}\left(\nabla_{k,t-L+\ell}\right)^{\intercal}(\wv_{t-L+1}-\wv_{\star})\nonumber\\
&\geq\sum_{k=1}^{K}\sum_{\ell=1}^{L}\Big[\Lc\left(f(\xv_{k,t-L+\ell};\wv_{t-L+1}),y_{k,t-L+\ell}\right)\nonumber\\
    &\quad\quad\quad\quad\quad\quad\quad\quad -\Lc\left(f(\xv_{k,t-L+\ell};\wv_{\star}),y_{k,t-L+\ell}\right)\Big]\nonumber\\
    &-\frac{3\beta\eta^2 L(L-1)}{1+6\beta^2\eta^2L(L-1)}\sum_{k=1}^{K}\sum_{\ell=1}^{L}\|\nabla_{k,t-L+\ell}(\wv_{\star})\|^2.\label{eq:lower1}
\end{align} Leveraging the upper-bound in \eqref{eq:up} and the lower-bound in \eqref{eq:lower1}, we have:
\begin{align}\label{eq:pf1_4}
&\sum_{k=1}^K\sum_{\ell=1}^L\Lc\left(f(\xv_{k,t-L+\ell};\wv_{t-L+1}),y_{k,t-L+\ell}\right)\nonumber\\
&\quad\quad\quad\quad\quad\quad -\sum_{k=1}^{K}\sum_{\ell=1}^{L}\Lc\left(f(\xv_{k,t-L+\ell};\wv_{\star}),y_{k,t-L+\ell}\right)\nonumber\\
&\leq \frac{K}{2\eta}\left(\|\wv_{t-L+1}-\wv_{\star}\|^2-\EE_{t+1}\left[\|\wv_{t+1}-\wv_{\star}\|^2\right]\right)\nonumber\\
&+3\beta\eta^2L(L-1)\sum_{k=1}^{K}\sum_{\ell=1}^{L}\|\nabla_{k,t-L+\ell}(\wv_{\star})\|^2\nonumber\\
    &+\frac{2 \eta L }{p_{\rm min}}\left[1+\frac{\sigma_{\rm q}^2(p_{\rm sum}-p_{\rm min}+1)}{K}\right]\sum_{k=1}^{K}\sum_{\ell=1}^{L}\|\nabla_{k,t-L+\ell}(\wv_{\star})\|^2,
\end{align}where we used the fact that $(1-4\eta \beta L R/p_{\rm min})/(1+6\beta^2\eta^2 L(L-1))<1$.
In fact, this approximation is quite accurate when $T$ is sufficiently large as $\eta=\Oc(1/\sqrt{T})$. Taking the conditional expectations in the both sides, we can get:
\begin{align}\label{eq:17}
&\sum_{k=1}^K\sum_{\ell=1}^L\Big[\EE_{t}\left[\Lc\left(f(\xv_{k,t-L+\ell};\wv_{t-L+1}),y_{k,t-L+\ell}\right)\right]\nonumber\\
    &\quad\quad\quad\quad\quad\quad -\Lc\left(f(\xv_{k,t-L+\ell};\wv_{\star}),y_{k,t-L+\ell}\right)\Big]\nonumber\\
    &\leq\frac{K}{2\eta}\left(\EE_{t}\left[\|\wv_{t-L+1}-\wv_{\star}\|^2\right]-\EE_{t+1}\left[\|\wv_{t+1}-\wv_{\star}\|^2\right]\right)\nonumber\\
&+3\beta\eta^2L(L-1)\sum_{k=1}^{K}\sum_{\ell=1}^{L}\|\nabla_{k,t-L+\ell}(\wv_{\star})\|^2\\
    &+\frac{2 \eta L }{p_{\rm min}}\left[1+\frac{\sigma_{\rm q}^2(p_{\rm sum}-p_{\rm min}+1)}{K}\right]\sum_{k=1}^{K}\sum_{\ell=1}^{L}\|\nabla_{k,t-L+\ell}(\wv_{\star})\|^2,\nonumber
\end{align}
where we used the fact that 
\begin{align*}
    &\EE\left[\|\wv_{t+1}-\wv_{\star}\|^2|\wv_{t-2L+1},\wv_{t-L+1}\right]\nonumber\\
    &\quad=\EE\left[\|\wv_{t+1}-\wv_{\star}\|^2|\wv_{t-L+1}\right]=\EE_{t+1}\left[\|\wv_{t+1}-\wv_{\star}\|^2\right],
\end{align*} namely, $\wv_{t+1}$ is conditionally independent from $\wv_{t-2L+1}$ given $\wv_{t-L+1}$. Note that \eqref{eq:17} implies the upper bound on the cumulative loss of the $L$ consecutive time slots (i.e., time slots $t-L+1, t-L+2,...,t$ for some $t \in [T]_L$). Summing \eqref{eq:17} over all $t \in [T]_L$ and by telescoping sum, we have:
\begin{align}\label{eq:18}
&\sum_{t=1}^T\sum_{k=1}^K \Big[\EE_{t}\left[\Lc\left(f(\xv_{k,t};\wv_{\psi_L(t)}),y_{k,t}\right)\right]\nonumber\\
&\quad\quad\quad\quad\quad\quad\quad\quad\quad\quad\quad\quad\quad\quad -\Lc\left(f(\xv_{k,t};\wv_{\star}),y_{k,t}\right)\Big]\nonumber\\
    &\stackrel{(a)}{\leq}\frac{K}{2\eta}\left(\|\wv_{\star}\|^2-\EE_{T+1}\left[\|\wv_{T+1}-\wv_{\star}\|^2\right]\right)\nonumber\\
    &\quad+\frac{2 \eta L K T \sigma_{\rm diff}^2}{p_{\rm min}}\left[1+\frac{\sigma_{\rm q}^2(p_{\rm sum}-p_{\rm min}+1)}{K}\right]\nonumber\\
    &\quad+3\beta \eta^2 L^2 K  T \sigma_{\rm diff}^2\nonumber\\
    &\stackrel{(a)}{\leq}\frac{K\|\wv_{\star}\|^2}{2\eta}+\frac{2 \eta L K T \sigma_{\rm diff}^2}{p_{\rm min}}\left[1+\frac{\sigma_{\rm q}^2(p_{\rm sum}-p_{\rm min}+1)}{K}\right]\nonumber\\
    &\quad+3\beta \eta^2 L(L-1) K  T \sigma_{\rm diff}^2,\nonumber
\end{align}
where (a) is from Definition 1 and (b) is due to the fact that $\|\wv_{T+1}-\wv_{\star}\|^2\geq0$. This completes the proof of Theorem 2.

\begin{lemma} The following inequality holds:
\begin{align}
&\sum_{k=1}^{K}\sum_{\ell=1}^{L}\|\wv_{t-L+1} - \gv_{k,t-L+\ell}\|^2 \nonumber\\
&\quad\quad\quad\leq \frac{3\beta \eta^2 L(L-1)}{2}\sum_{k=1}^{K}\sum_{\ell=1}^{L}\|\nabla_{k,t-L+\ell}\|^2.
\end{align}
\end{lemma}
\begin{IEEEproof} Note that when $L=1$ or $\ell=1$, we have that $\|\wv_{t-L+1} - \gv_{k,t-L+1}\|^2 = 0$. Assuming $L>1$, we have:
\begin{align}
    &\sum_{\ell=1}^{L}\left\Vert\wv_{t-L+1}-\gv_{k,t-L+\ell}\right\Vert^2=\sum_{\ell=2}^{L}\left\Vert\wv_{t-L+1}-\gv_{k,t-L+\ell}\right\Vert^2\nonumber\\
    &\quad =\sum_{\ell=2}^{L}\left\Vert\wv_{t-L+1}-\left(\gv_{k,t-L+\ell-1}-\eta\nabla_{k,t-L+\ell-1}\right)\right\Vert^2\nonumber
    \end{align}
    \begin{align}
    &\quad\stackrel{(a)}{\leq}\sum_{\ell=2}^{L}\left(1+\frac{1}{L-1}\right)\left\Vert\wv_{t-L+1}-\gv_{k,t-L+\ell-1}\right\Vert^2\nonumber\\
    &\quad\quad +\eta^2 L\sum_{\ell=2}^{L}\left\Vert\nabla_{k,t-L+\ell-1}\right\Vert^2\nonumber\\
    &\quad\stackrel{(b)}{\leq}\sum_{\ell=2}^{L}\sum_{\tau=0}^{\ell-2}\eta^2 L\left\Vert\nabla_{k,t-L+\ell-1-\tau}\right\Vert^2\left(1+\frac{1}{L-1}\right)^{\tau}\nonumber\\
    &\quad\leq \eta^2 L \left(1+\frac{1}{L-1}\right)^{L-1} \sum_{\ell=2}^{L}\sum_{\tau=0}^{\ell-2}\left\Vert\nabla_{k,t-L+\ell-1-\tau}\right\Vert^2\nonumber\\
    &\quad\stackrel{(c)}{\leq} 3 \eta^2 L(L-1) \sum_{\ell=1}^{L}\|\nabla_{k,t-L+\ell}\|^2 ,\label{eq:12}
\end{align} where (a) is from Lemma 1 with $a=\frac{1}{L-1}$, $(b)$ is from Lemma 4, and (c) is due to the fact that 
\begin{align*}
    \left(1+1/(L-1)\right)^{L-1}\leq \lim_{L\rightarrow\infty}\left(1+1/(L-1)\right)^{L-1}=e<3,
\end{align*} and for $L>1$,
\begin{align*}
\sum_{\ell=2}^{L}\sum_{\tau=0}^{\ell-2}\left\Vert\nabla_{k,t-L+\ell-1-\tau}\right\Vert^2 \leq (L-1)\sum_{\ell=1}^{L}\|\nabla_{k,t-L+\ell}\|^2.
\end{align*} This completes the proof.
\end{IEEEproof}

\begin{lemma}\label{lem3} For $\ell\geq 2$, the following inequality holds:
\begin{align}
 A\|\wv_t-\gv_{k,t+\ell-2}\|^2+B\|\nabla_{k,t+\ell-2}\|^2\nonumber\\
    \leq\sum_{\tau=0}^{\ell-2}B\|\nabla_{k,t+\ell-2-\tau}\|^2 A^{\tau},\label{eq:inq}
\end{align} where $A=(1+1/(L-1))$ and $B=L\eta^2$.
\end{lemma}
\begin{IEEEproof} The proof will be done by induction. When $\ell=2$, the inequality in \eqref{eq:inq} is trivial because $\wv_t = \gv_{k,t}$ and accordingly we have:
\begin{equation}
A\|\wv_t-\gv_{k,t}\|^2+B\|\nabla_{k,t}\|^2 =B\|\nabla_{k,t}\|^2.
\end{equation} Suppose that the inequality in  \eqref{eq:inq} holds for $ 2 \leq m < \ell$, i.e.,
\begin{align}
    &A\|\wv_t-\gv_{k,t+m-2}\|^2+B\|\nabla_{k,t+m-2}\|^2\nonumber\\
    &\leq \sum_{\tau=0}^{m-2}B\|\nabla_{k,t+m-2-\tau}\|^2 A^{\tau}.\label{eq:hypo}
\end{align} Focusing on $m+1$, then, we have:
\begin{align*}
    &A\|\wv_t-\gv_{k,t+m-1}\|^2+B\|\nabla_{k,t+m-1}\|^2\\
    &\stackrel{(a)}{\leq} A\left(A\|\wv_t-\gv_{k,t+m-2}\|^2+\|\nabla_{k,t+m-2}\|^2\right)\nonumber\\
    &\quad\quad + B\|\nabla_{k,t+m-1}\|^2\nonumber\\
    &\stackrel{(b)}{\leq} A\left(\sum_{\tau=0}^{m-2}B\|\nabla_{k,t+m-2-\tau}\|^2 A^{\tau}\right) + B\|\nabla_{k,t+m-1}\|^2\nonumber\\
    &=\sum_{\tau=0}^{(m+1)-2}B\|\nabla_{k,t+(m+1)-2-\tau}\|^2 A^{\tau}
\end{align*}where (a) is from Lemma~\ref{lem1} with $a=\frac{1}{L-1}$ and (b) is due to the hypothesis assumption in \eqref{eq:hypo}. This completes the proof.
\end{IEEEproof}


\section{Experiments}\label{sec:Exp}

Via experiments with real datasets, we verify the analytical results in  Section~\ref{sec:TA} and demonstrate the superiority of the optimized OFedIQ compared with the state-of-the-art OFL methods. Recall that using the stochastic quantizer in Definition 2, the proposed method is denoted by OFedIQ$(L,p,s,b)$, where $L$ is the transmission period, $p$ is the sampling rate, and $(s,b)$ is the parameter of the quantization. In convention, when quantization is not applied, it is denoted as OFedIQ$(L,p,\infty)$. For comparison, the following benchmark methods are considered:
\begin{itemize}[leftmargin=*]
\item {\bf FedOGD} \cite{hong2021communication}: This is the counterpart of FedSGD in FL. It is considered as the performance-limit of communication-efficient OFL methods.
\item {\bf OFedAvg($p$)}: This is the counterpart of FedAvg \cite{mcmahan2017communication}. In this method, a smaller number of clients are participated at every time step (i.e., $pK$ in average).
\item {\bf FedOMD($L$)}\cite{mitra2021online}: This is the communication-efficient OFL method based on OMD and periodic transmission. In our experiments, a commonly adopted Euclidean distance is used as the underlying Bregman divergence of OMD.
\end{itemize} To the best of our knowledge, there is no other OFL method with a performance guarantee. In fact, OFedIQ$(L,p,s,b)$ subsumes the  benchmark methods as special cases. Nonetheless, it does not directly imply that OFedIQ can outperform them. Namely, the parameters should be optimized, which is efficiently performed via the proposed Algorithm~\ref{alg:parameter-opt} 
For a hyperparameter $\eta$, it was proved in Section~\ref{sec:TA} that $\eta=\Oc(1/\sqrt{T})$ is asymptotically optimal for the proposed and benchmark methods. In practice, however, it is generally unavailable to see the end-point of continuous streaming data (i.e., $T$) in advance. We set $\eta=0.01$ for all experiments, which is sufficient for their fair comparisons. 
Let $\hat{y}_{k,\tau}$ and $y_{k,\tau}$ denote a predicted label and a true label, respectively. As commonly adopted in online learning tasks \cite{shen2019random,hong2020active, hong2021communication}, we use the following metric to assess learning accuracy:
\begin{itemize}[leftmargin=*]
\item Online classifications: 
\begin{equation*}
{\rm Accuracy}(t) =1- \frac{1}{tK}\sum_{\tau=1}^{t}\sum_{k=1}^{K} \min\{1, |\hat{y}_{k,\tau} -y_{k,\tau}| \}.
\end{equation*}
\item Online regressions: 
\begin{equation*}
        {\rm MSE}(t)=\frac{1}{tK}\sum_{\tau=1}^{t}\sum_{k=1}^{K} (\hat{y}_{k,\tau} - y_{k,\tau})^2, 
\end{equation*}  where in the experiments, labels $y_{k,t}$'s are normalized as  $y_{k,t} \in [0,1]$ for $\forall k \in [K]$ and $\forall t \in [T]$.
\end{itemize} To evaluate the communication-cost reduction (CCR) of the communication-efficient (CF) methods (OFedAvg, FedOMD, and OFedIQ) over the baseline method (FedOGD), the following metric is considered:
\begin{align*}
    \mbox{CCR}(\%)
  &=(1-\gamma)\times 100,
\end{align*} where $\gamma=\frac{\mbox{communication-cost of CF-method}}{32KD}$ as introduced in Section~\ref{subsec:opt}.
One of the following model architectures will be adopted according to dataset:
\begin{itemize}[leftmargin=*]
\item{{\bf Model I (CNN):}} This convolutional neural network (CNN) model consists of two convolutional layers (each with $32$ nodes and $64$ nodes), one softmax layer with $10$ nodes, and $3$ bias nodes. Each convolutional layer is composed of multiple sequential operations including convolution, ReLU activation, and max-pooling. A cross-entropy loss function is used. This model contains the $109$ nodes and accordingly, the number of parameters is equal to
$D=10\times 32+289\times 64+1601\times 10 = 34,826$.
\item{\bf Model II (DNN):} This deep neural network (DNN) model consists of two ReLU layer with $64$ nodes, 3 bias nodes, and one softmax layer with $4$ nodes. A cross-entropy loss function is considered.  This model contains the overall $135$ nodes and accordingly, the number of parameters is equal to $D=2\times 64+65\times 64+65\times 4 = 4,548$.
\item{{\bf Model III (DNN):}} This deep neural network (DNN) model consists of two ReLU layer with $64$ nodes, one fc layer with $1$ node, and $3$ bias nodes. A regularized least-square loss function $\Lc(\cdot,\cdot)$ is used. This model contains the $132$ nodes and accordingly, the number of parameters is equal to $D=2\times 64+65\times 64+65\times 1=4,353$.
\end{itemize}



\noindent In our experiments, the network has $K=1000$ clients each receiving $T$ data samples sequentially. The following real datasets are considered:
\begin{itemize}[leftmargin=*]
\item {\bf MNIST \cite{lecun1998gradient}:} The dataset consists of 60,000 samples with 10 handwritten single digits. Each sample has 784 features. The goal is to classify a given handwritten digit into one of 10 classes (integer values from 0 to 9). Model I is used.
\item \textbf{Room Occupancy Estimation \cite{singh2018machine}:} This dataset contains 10,130 time-series samples with the four classes $\{0,1,2,3\}$. Each sample has 14 features extracted from environmental sensors such as temperature, light, sound, CO2 and PIR. The goal is to predict the precise number of occupants in a room. Model II is used.
\item \textbf{Air quality \cite{SDeVito2008}:} This dataset contains 38,563 time-series samples. Each sample has 5 features including hourly response from an array of 5 metal oxide chemical sensors embedded in a city of Italy.  The goal is to estimate the concentration of polluting chemicals in the air. Model III is used.

\item \textbf{Appliances energy \cite{Candanedo2017}:} This dataset contains 18,604 samples, each of which has $25$ features describing appliances energy uses such as temperature and pressure in houses. The goal is to predict energy use in a low-energy building. Model III is used.
\end{itemize}
For the case of $N_b < 1000 T$, where $N_b$ denotes the number of whole data samples in a dataset, the entire samples are repeated by $\lfloor 1000T/N_b \rfloor$ times. To study OFL, it is required to specify how data samples in each dataset are distributed across the $1000$ clients. 
As in \cite{mcmahan2017communication}, random partitioning is performed, from which the entire $1000 T$ data samples are shuffled and then partitioned into the $K=1000$ clients. The resulting partition is denoted as $\{\Dc_{1},...,\Dc_{K}\}$ with $|\Dc_{k}|=T, \forall k \in [K]$.
At time step $t$, each client $k$ receives the data $(\xv_{k,t}, y_{k,t})$ from $\Dc_k$ sequentially. Although there are numerous partitions, our experiments confirmed that the performances of the proposed and benchmark methods are not affected by partitions as long as $T$ is large. This coincides with our theoretical results as partitions do not change the degree of data-heterogeneity $\sigma_{\rm diff}$.

\begin{figure}[!t]
\centerline{\includegraphics[width=8.0cm]{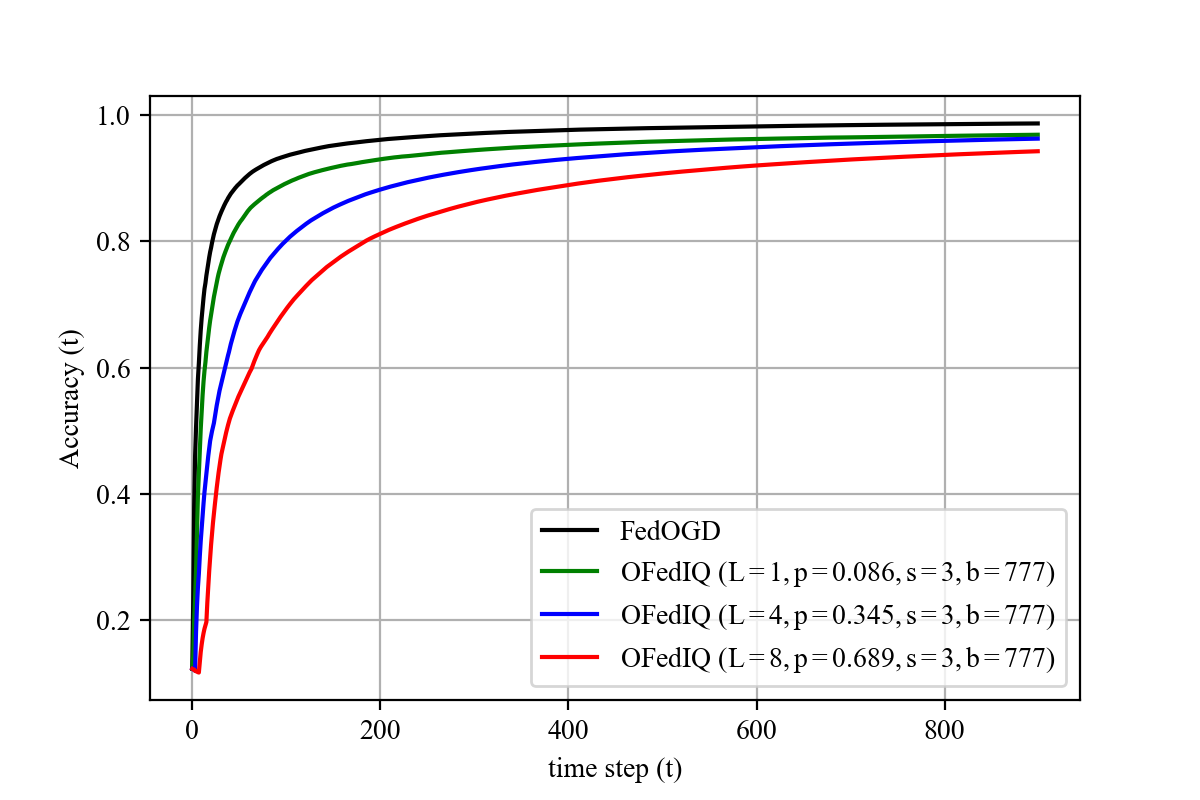}}
\caption{Performances of OFedIQ according to the choices of transmission period $L$, where $K=1000$ and $\mbox{CCR}=99\%$.}
\label{fig3}
\end{figure}
\begin{figure}[!t]
\centerline{\includegraphics[width=8.0cm]{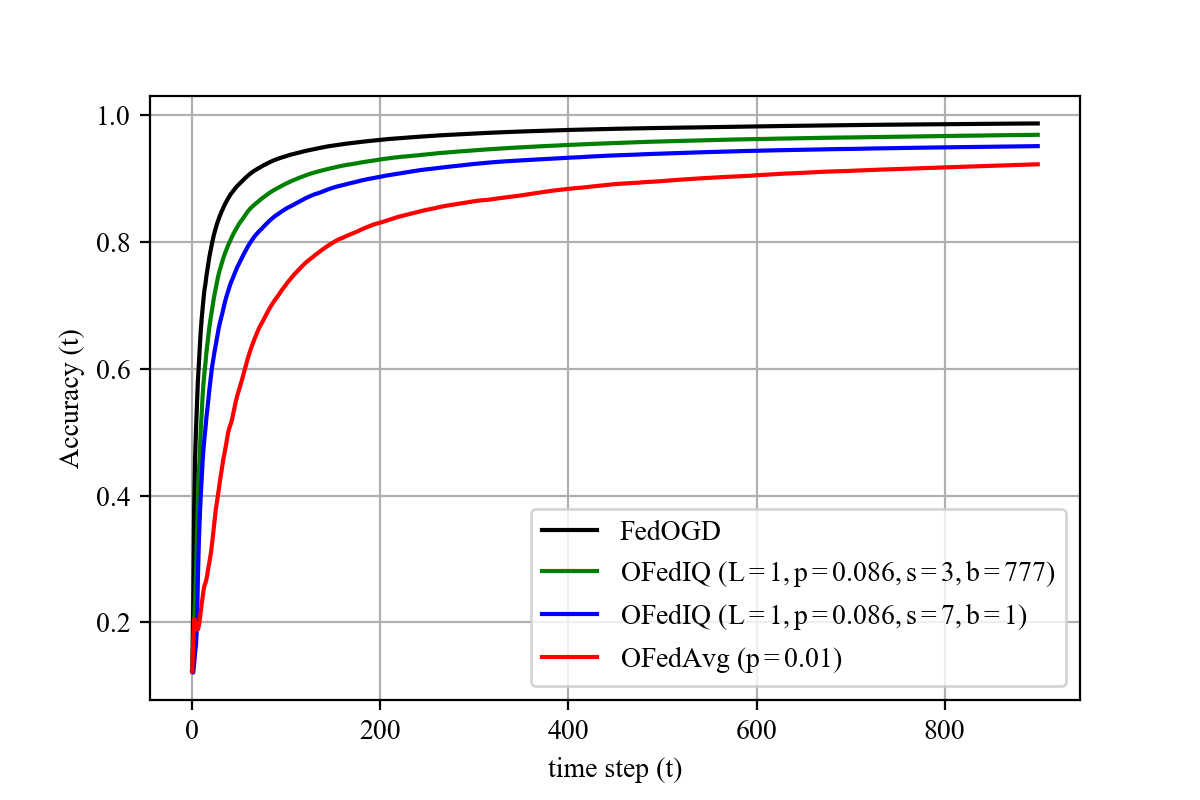}}
\caption{Performance improvements of OFedIQ due to the proposed optimized quantizer, where $K=1000$ and $\mbox{CCR}=99\%$.}
\label{fig4}
\end{figure}

\begin{figure*}[!h]
    \centering
    \subfigure[MNIST ($\mbox{CCR}=90\%$)]{
        \includegraphics[width=0.47\linewidth]{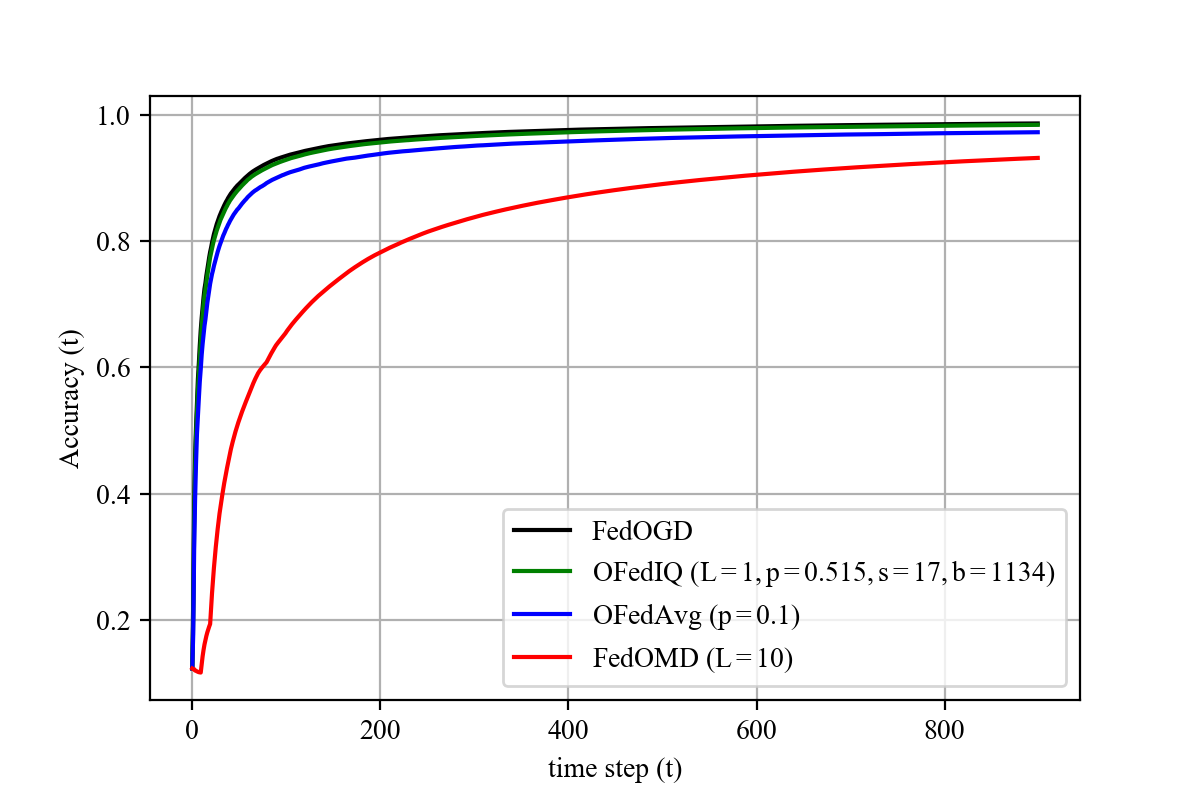}
    }
    \centering
    \subfigure[MNIST ($\mbox{CCR}=99\%$)]{
        \includegraphics[width=0.47\linewidth]{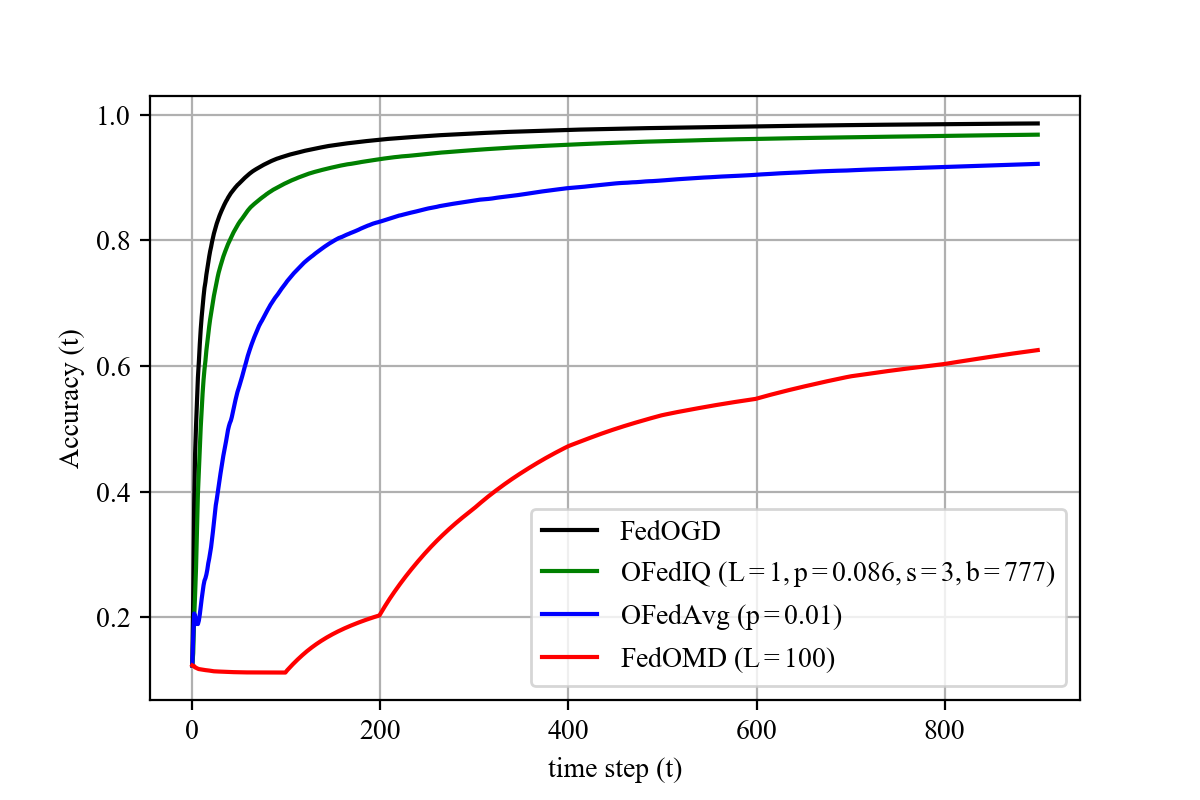}
    }
    \centering
    \subfigure[Air Quality ($\mbox{CCR}=90\%$)]{
        \includegraphics[width=0.47\linewidth]{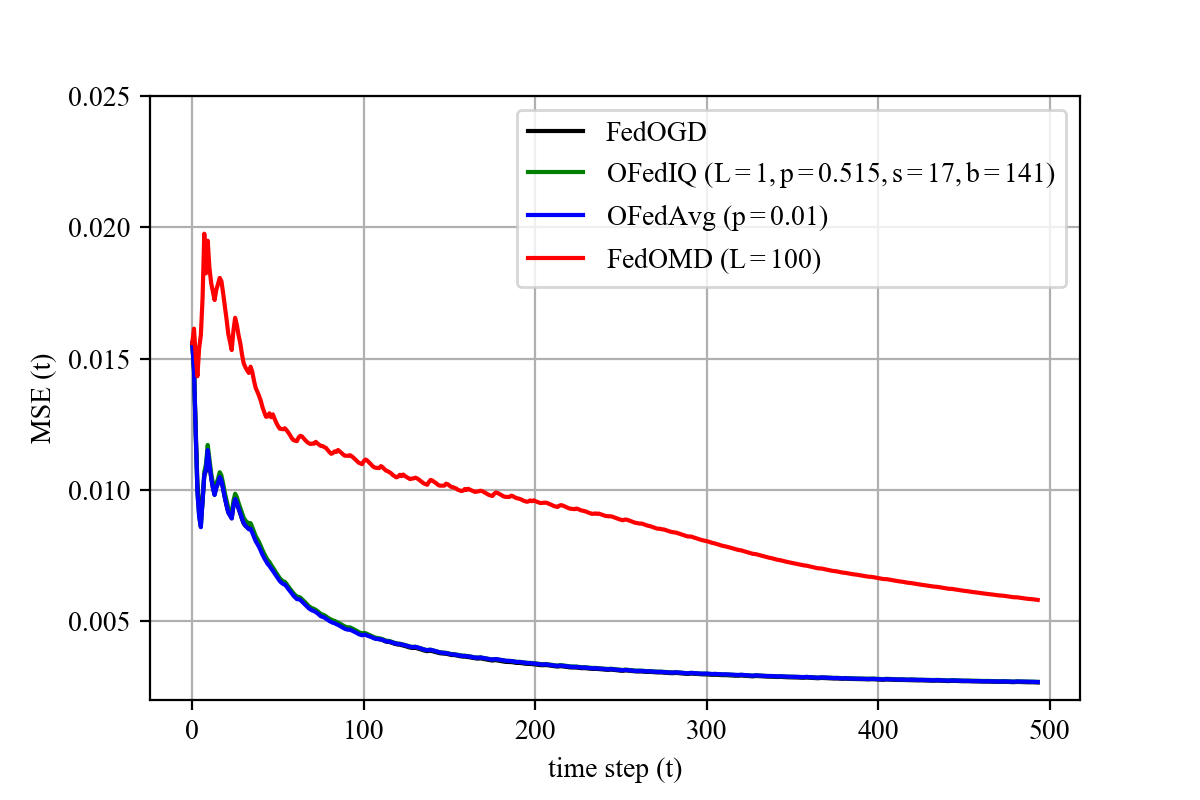}
    }
    \centering
    \subfigure[Air Quality ($\mbox{CCR}=99\%$)]{
        \includegraphics[width=0.47\linewidth]{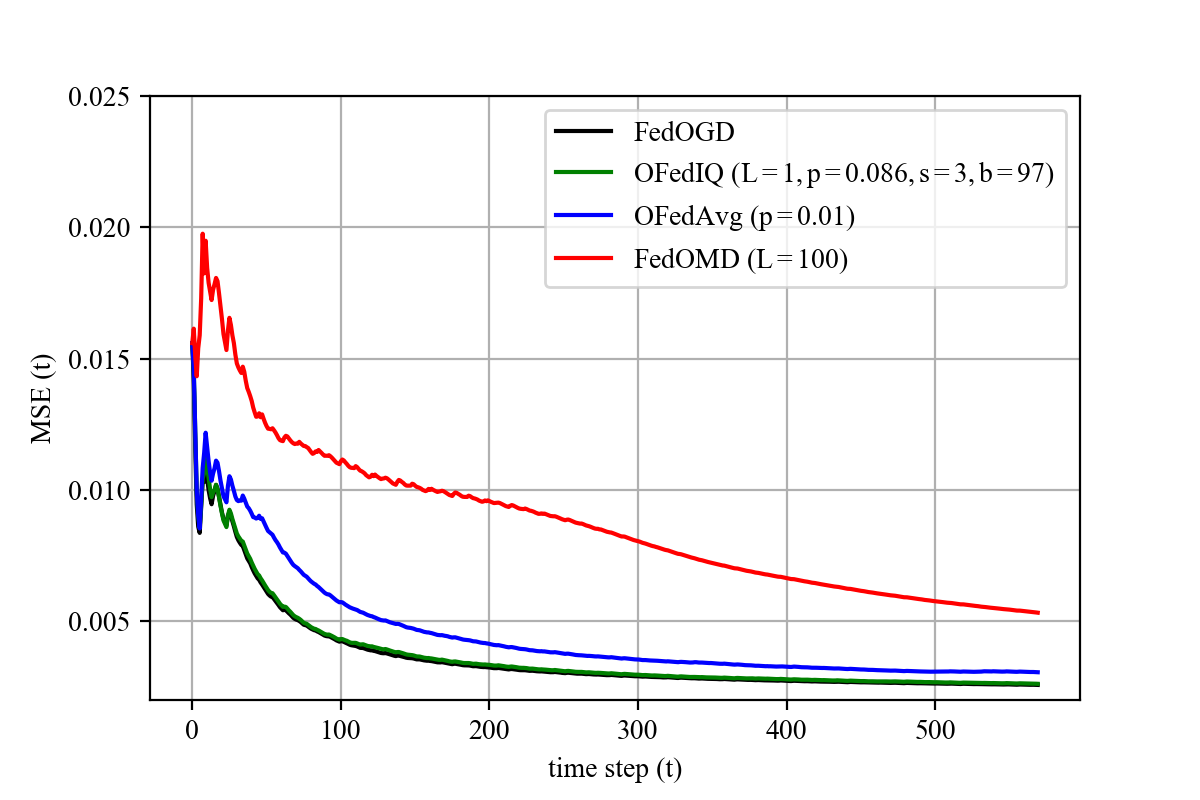}
    }
    \centering
    \subfigure[Room Occupancy Estimation ($\mbox{CCR}=99\%$)]{
        \includegraphics[width=0.47\linewidth]{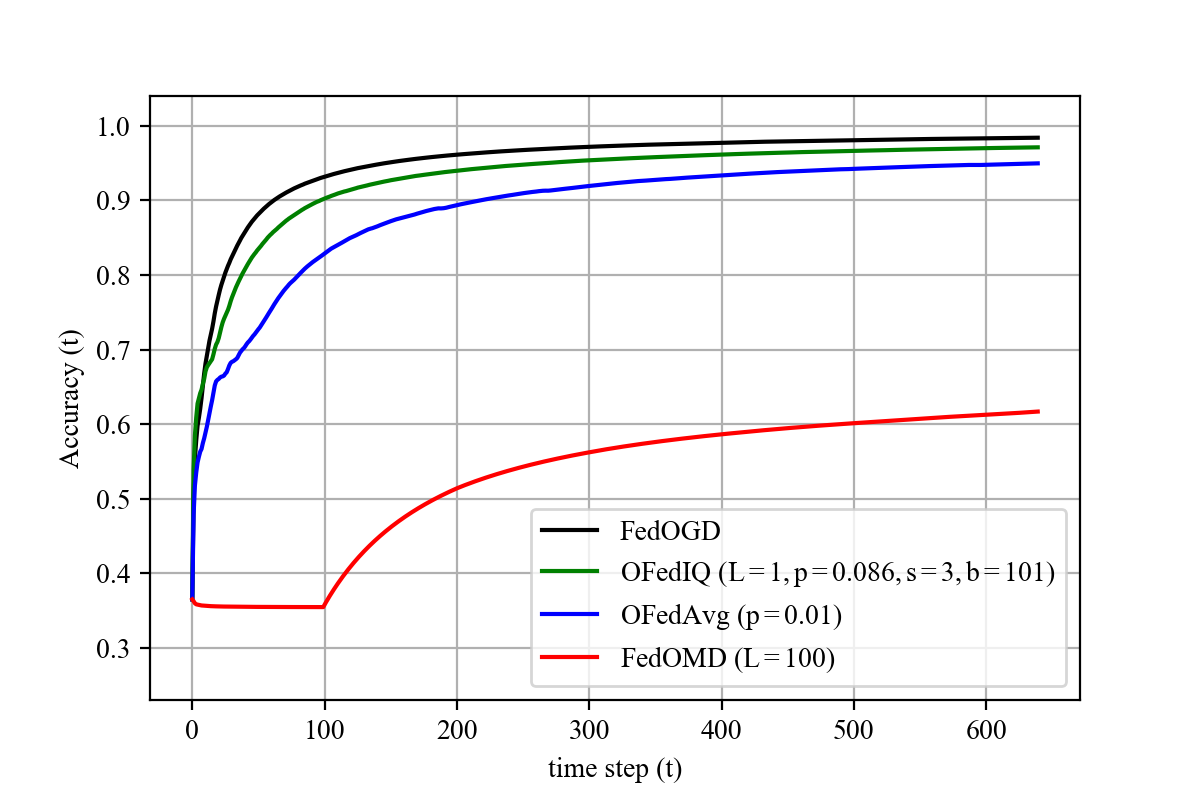}
    }
    \centering
    \subfigure[Appliances Energy ($\mbox{CCR}=99\%$)]{
        \includegraphics[width=0.47\linewidth]{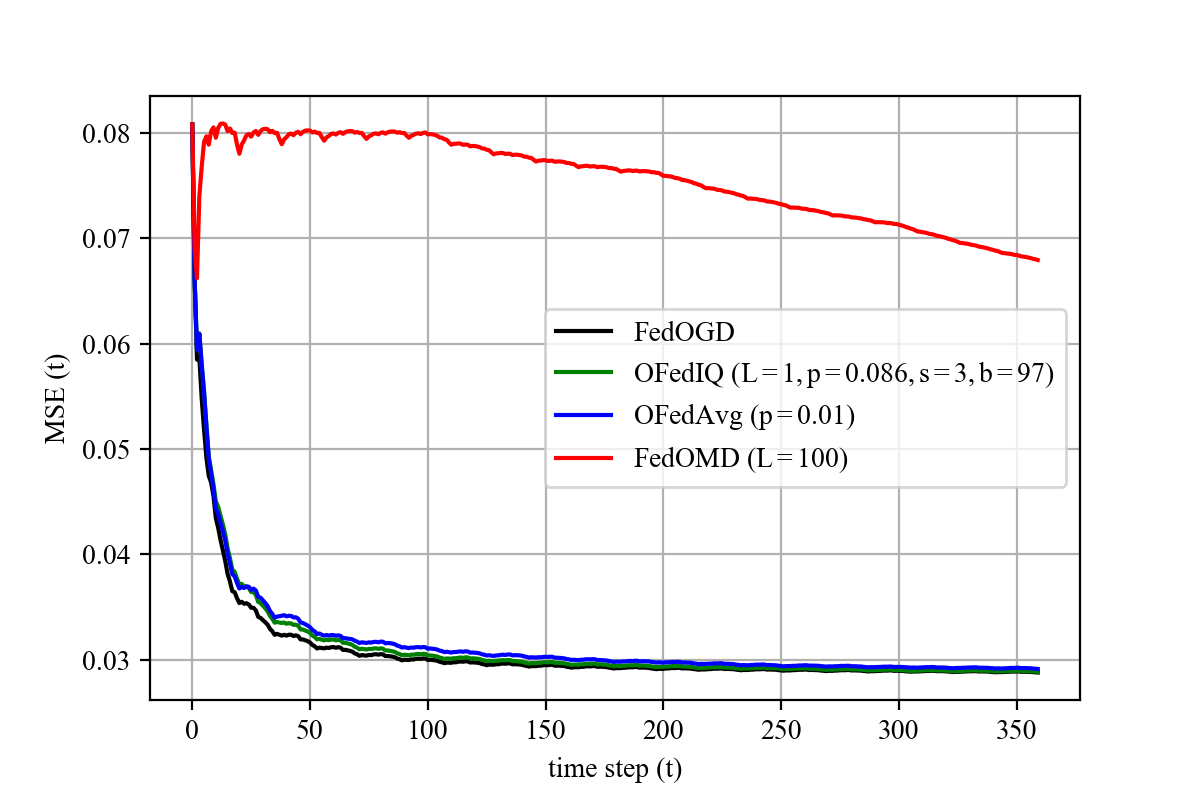}
    }
    \caption{Performance comparisons of various OFL methods for online classification and regression tasks, where $K=1000$. The green lines depict the performances of the {\bf optimized OFedIQs}. In (c), FedOGD, OFedIQ, and OFedAvg yield the same performance.}
\end{figure*}

\vspace{0.1cm}
\noindent{\bf Performance evaluation:} We first validate the analytical results in Section~\ref{sec:TA} via experiments with MNIST data. Fig.~\ref{fig3} manifests that $L=1$ is the best choice for a fixed $p/L$, which is well-matched with our analysis in Section~\ref{subsubsec:pa}. Specifically, the performance loss with $L > 1$ is noticeable in non-asymptotic case (e.g., $t=200$), whereas it becomes negligible as $t$ is sufficiently large. Also, this loss becomes severer as $L$ increases. Based on these results and our analysis in Section~\ref{subsubsec:pa}, to reduce the communication costs as intermittent transmission, client subsampling should be adopted rather than periodic transmission. Fig~\ref{fig4} shows the efficacy of the proposed parameter optimization in Algorithm 3. The optimized OFedIQ$(1,0.086,3,777)$ outperforms OFedIQ with different parameters such as OFedIQ$(1,0.086,7,1)$ and OFedIQ$(1,0.01,\infty)$ (a.k.a. OFedAvg$(0.01)$), while having the same communication costs. To the best of our efforts, we could not find other OFedIQ showing better performance than the optimized OFedIQ. Compared with OFedIQ$(1,0.086,7,1)$, the optimized OFedIQ used the proposed stochastic quantization in Definition 2 rather than the stochastic quantization in \cite{alistarh2017qsgd, reisizadeh2020fedpaq}. From Fig.~\ref{fig4}, we can identify the effectiveness of the proposed stochastic quantizer. Moreover, the comparison with FedAvg demonstrates that quantization is indeed useful in order to reduce the communication costs considerably. This supports our theoretical analysis in Section~\ref{subsubsec:op}.

Fig. 4 shows the performances of various OFL methods on regression and classification tasks with real datasets. As aforementioned, FedOGD is considered as the performance bound of the communication efficient OFL methods. Regarding the optimized OFedIQ$(L,p,s,b)$, $L=1$ is chosen from Section~\ref{subsubsec:pa}, and $p$ and $(s,b)$ are optimized via Algorithm 3. From Fig. 4, we observe that the optimized OFedIQ outperform the benchmark methods for all datasets. Remarkably, it can almost achieve the performance of FedOGD while reducing the communication costs by $99\%$. Interestingly, FedOMD, which reduces the communication costs via periodic transmission only, yields very poor performances especially when $\mbox{CCR}=99\%$. This is because only a small number of global updates are performed. These results strengthen the argument in Section~\ref{subsubsec:pa} that periodic transmission is not suitable for OFL. When $\mbox{CCR}=90\%$, it seems that quantization is unnecessary since FedAvg has the same performance as the optimized OFedIQ. To further reduce the communication costs (e.g., $\mbox{CCR}=99\%$), however, the optimized quantization is indeed effective. Thus, the optimized OFedIQ can provide a more elegant tradeoff of performance and communication costs than FedAvg. Our experimental results suggest the practicality of the optimized OFedIQ.

\section{Conclusion}\label{sec:con}
We presented a communication-efficient method (named OFedIQ) for online federated learning (OFL) by means of client subsampling, periodic transmission, and quantization. Through our tighter regret analysis, 
it was proved that OFedIQ can enjoy an optimal sublinear regret for any real dataset (up to a constant) while significantly reducing the communication costs of FedOGD (i.e., the performance-limit). On the basis of our analysis, moreover, we efficiently optimized the key parameters of OFedIQ such as the sampling rate, the transmission period, and quantization levels. Via experiments with real datasets, we demonstrated the effectiveness of our algorithm on various online classification and regression tasks. Specifically, for all datasets in our experiments, the optimized OFedIQ outperforms the state-of-the-art OFL methods. In addition, it can almost achieve the performances of FedOGD while reducing the communication costs by $99\%$. These results suggest the practicality of the optimized OFedIQ.

\ifCLASSOPTIONcompsoc
  \section*{Acknowledgments}
\else
  \section*{Acknowledgment}
\fi

This work was supported by the National Research Foundation of Korea (NRF) grant funded by the Korea government (MSIT) (NRF-2020R1A2C1099836).

\ifCLASSOPTIONcaptionsoff
  \newpage
\fi

\bibliographystyle{IEEEtran}
\bibliography{example_paper}

\end{document}